\documentclass{article}
\PassOptionsToPackage{dvipsnames,table}{xcolor}
\usepackage{arxiv}
\usepackage{graphicx}
\usepackage[numbers]{natbib}
\usepackage{doi}
\usepackage[utf8]{inputenc} 
\usepackage[T1]{fontenc}    
\usepackage{url}            
\usepackage{booktabs}       
\usepackage{amsfonts}       
\usepackage{nicefrac}       
\usepackage{microtype}      
\usepackage{amsmath}
\usepackage{amssymb}
\usepackage{bbm}
\usepackage{wasysym}
\usepackage{pifont}
\usepackage{threeparttable}
\usepackage{subcaption}
\usepackage{array}
\usepackage{makecell}
\usepackage{algpseudocode} 
\usepackage{multirow}
\usepackage{listings}
\usepackage{enumitem}
\usepackage{hyperref}
\usepackage{todonotes}
\usepackage{comment}
\usepackage[capitalise, noabbrev]{cleveref}
\usepackage{lscape}
\usepackage{tabularray}
\usepackage{CJK}

\usepackage{colortbl}
\usepackage{wrapfig}
\usepackage{adjustbox}
\usepackage{verbatim}
\usepackage{minitoc}
\usepackage{algorithm}
\newcommand{\xmark}{\ding{55}}

\definecolor{bestgreen}{RGB}{208,124,176}
\definecolor{secondgreen}{RGB}{235,170,210}
\definecolor{thirdgreen}{RGB}{255,225,245}
\definecolor{bestpurple}{RGB}{120,70,255}

\title{VGGT-Prime: Compute-Adaptive Mixture-of-Heads for Efficient Visual Geometry Transformers}

\author{%
  Abteen Arab$^{1,2,*}$ \quad Guile Wu$^{1}$ \quad Chengjie Huang$^1$ \quad Dongfeng Bai$^1$\\
  $^{1}$Huawei Noah's Ark Lab \quad $^{2}$University of British Columbia\\
  \texttt{aarab02@student.ubc.ca, guile.wu@outlook.com, chengjie.huang@outlook.com} \\
  \texttt{baidongfeng@huawei.com} \\
}

\renewcommand{\headeright}{Technical Report}

\begin{document}

\maketitle

\begin{figure}[h]
    \centering
    \includegraphics[width=0.99\textwidth]{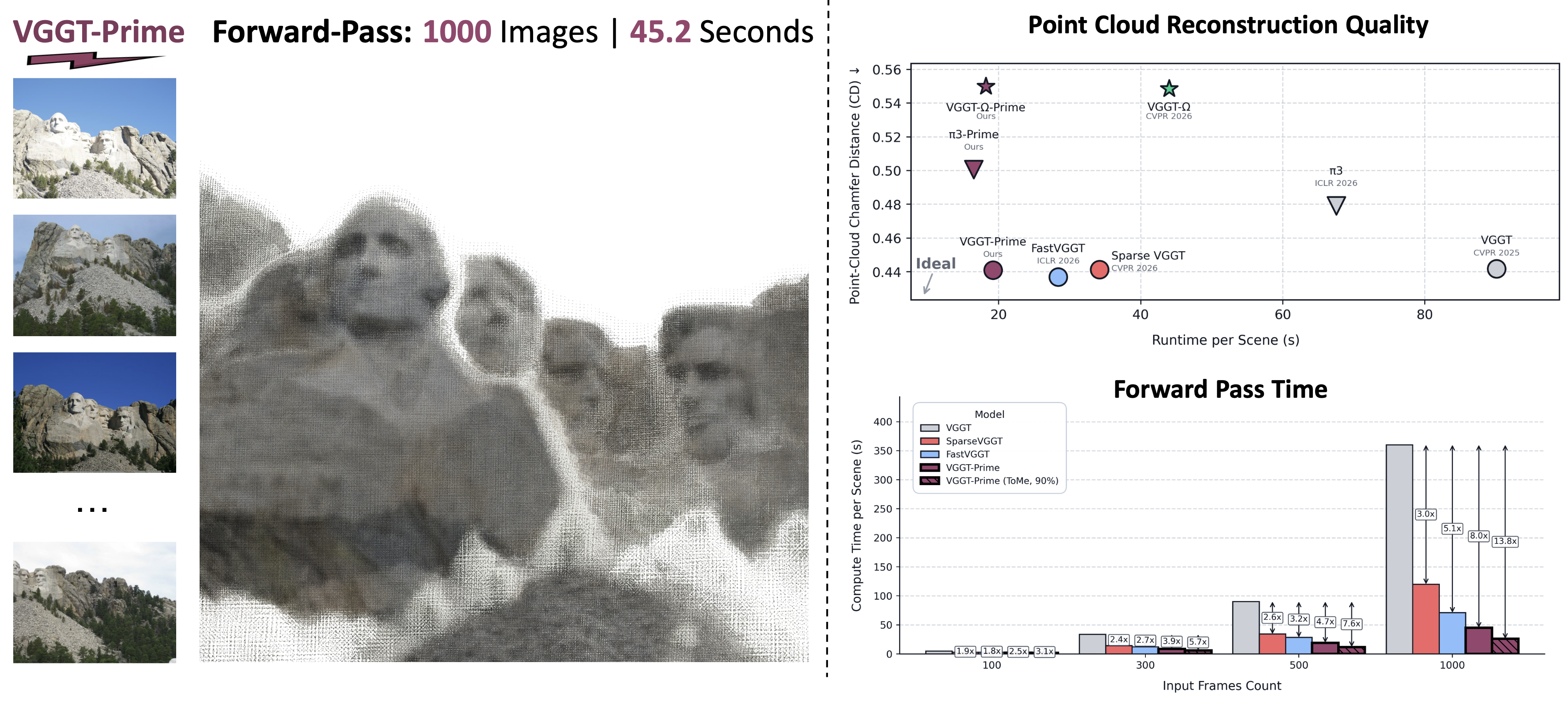}\par
    \captionof{figure}{
    VGGT-Prime delivers competitive ScanNet-500 reconstruction across backbones (top-right) and fast long-sequence inference, further accelerated by token merging (bottom-right).
    }
    \label{fig:main}
\end{figure}

\begingroup
\renewcommand\thefootnote{*}
\footnotetext{Abteen Arab contributed to this work during an internship at Huawei Canada.}
\endgroup

\thispagestyle{fancy}

\begin{abstract}
Feed-forward visual geometry models such as the Visual Geometry Grounded Transformer (VGGT) have recently enabled direct 3D reconstruction from multi-view images.
Despite their promising performance, these models scale quadratically with the number of input views due to their global attention mechanism, resulting in substantial latency for long sequence inputs.
There have been some recent efforts to accelerate VGGT, but they primarily focus on reducing \emph{token redundancy} through token merging or key/value sparsification.
Our work resolves this bottleneck from a different perspective by investigating \emph{architectural redundancy} in visual geometry transformers.
We show that the multi-head attention modules in VGGT’s global-attention layers contain substantial architectural redundancy, with only a subset of heads carrying critical geometric information.
In light of this observation, we propose VGGT-Prime, a compute-adaptive mixture-of-heads model that resolves this redundancy to accelerate visual geometry transformers while maintaining competitive reconstruction quality.
The key idea of VGGT-Prime is to estimate the appropriate computation level for each global-attention head using a lightweight router and then dynamically assign each head to different computation modes.
Extensive experiments on multiple datasets demonstrate that VGGT-Prime can achieve an {$8\times$} inference speedup over VGGT while maintaining competitive performance on camera pose, depth, and point-cloud predictions.
We further show that VGGT-Prime is complementary to existing acceleration methods, such as token merging, further improving inference speed by up to $14{\times}$ over VGGT.
An overview of our work is available on our \href{https://vggt-prime.github.io}{project page}.

\end{abstract}

\section{Introduction}

Reconstructing 3D scene geometry from multi-view images has long been a fundamental problem in computer vision.
Traditional reconstruction pipelines, such as structure from motion \cite{schoenberger2016sfm} and multi-view stereo \cite{schoenberger2016mvs}, rely on explicit geometric modeling and iterative optimization.
This makes them computationally expensive and difficult to run at scale, while remaining less robust in complex real-world scenarios with sparse views and ambiguous scene geometry.

Recent work has shifted toward feed-forward reconstruction methods that directly reconstruct 3D scene geometry from 2D images without complex iterative optimization \cite{dust3r_cvpr24, wang2025vggt, keetha2026mapanything, wang2026pi}.
Among them, the Visual Geometry Grounded Transformer (VGGT) has emerged as a leading approach \cite{wang2025vggt}, which predicts 3D geometric attributes such as camera parameters, depth maps, point maps, and point tracks in a single forward pass.
This strong performance is partly enabled by a global attention mechanism that performs self-attention over all tokens from all views, allowing the model to capture cross-view correspondences.
However, the computational cost of global attention scales quadratically with the number of input views, limiting the scalability and efficiency of VGGT.

More recently, several works have been proposed to alleviate the computational cost of global attention in VGGT \cite{shen2026fastvggt,wang2025fastervggt,wang2026httm,huang2026turbovggt}.
These methods primarily address \emph{token redundancy} through token merging or attention sparsification.
However, they do not directly address inefficiencies within the global attention architecture itself.
In this work, we tackle this bottleneck from an architectural perspective by asking: \textit{can VGGT be made more efficient by exploiting redundancy in its global attention architecture?}

To answer this question, we conduct an in-depth analysis of head saliency and attention patterns of global attention heads in VGGT.
We first measure the saliency of each attention head by evaluating its contribution to the final prediction and find that many heads can be removed with little impact on reconstruction quality, suggesting that they are redundant.
This conclusion is further supported by an analysis of attention patterns, which shows that low-saliency heads exhibit a strong mean pooling behaviour, while high-saliency heads are more selective and view-specific.
This analysis suggests that VGGT’s global multi-head attention contains substantial \emph{architectural redundancy} and that many attention heads can be replaced with computationally efficient approximations.

In light of this observation, in this work, we propose VGGT-Prime, a compute-adaptive mixture-of-heads model for efficient feed-forward visual geometry reconstruction.
Specifically, VGGT-Prime resolves the \emph{architectural redundancy} of the global attention mechanism in visual geometry transformers from a head-level perspective.
Rather than applying full softmax attention uniformly across all global attention heads, VGGT-Prime first uses a lightweight router to adaptively estimate the compute required for each head.
Then, it routes each head to one of three computation modes, including mean pooling, linear complexity surrogate attention, or exact softmax attention. 
By adaptively selecting the computational complexity of each head, VGGT-Prime reduces the computational cost of global attention while maintaining reconstruction quality.
We conduct extensive experiments to validate the effectiveness of VGGT-Prime on seven 3D reconstruction benchmark datasets.
Our results show that VGGT-Prime achieves an {$8\times$} speedup over VGGT\footnote[1]{Throughout this study, we compare against an accelerated VGGT variant introduced by \citet{keetha2026mapanything}.} on long sequences with 1000 frames, while maintaining competitive reconstruction quality.
In addition, we demonstrate that VGGT-Prime is generalizable to different visual geometry transformer backbones \cite{depthanything3, wang2026pi} and is complementary to existing acceleration methods, such as token merging \cite{shen2026fastvggt}, for further acceleration.

Our main \textbf{contributions} are summarized as follows:
\begin{itemize}
    \item We present an in-depth analysis of the global attention mechanism in visual geometry transformers, revealing the presence of head-level architectural redundancy;
    \item We propose VGGT-Prime, a compute-adaptive mixture-of-heads model for efficient visual geometry transformers, which adaptively routes each global attention head to an appropriate computation mode to reduce computational cost while maintaining reconstruction quality;
    \item We conduct extensive experiments on several 3D reconstruction benchmark datasets, demonstrating that VGGT-Prime achieves state-of-the-art efficiency-accuracy trade-offs, generalizes across different backbones, and can be combined with existing acceleration methods to deliver further speedups.
\end{itemize}

\section{Related Work}

\paragraph{Feed-Forward Multi-View Reconstruction.}
Reconstructing 3D scene geometry from a collection of images is a long-standing problem in computer vision.
This task involves estimating camera poses, depth maps, point clouds, and other geometric quantities.
Classical approaches address this task through scene level optimization pipelines such as structure from motion \cite{schoenberger2016sfm} and multi view stereo~\cite{schoenberger2016mvs}.
Although these methods are effective, they require engineered features and expensive iterative optimization algorithms, which makes them impractical for many applications.
Recently, feed-forward models have shown that 3D geometry can be reconstructed directly from images without iterative optimization~\cite{dust3r_cvpr24,wang2025vggt,wang2026pi,keetha2026mapanything}.
DUSt3R introduces dense point-map prediction of image pairs \cite{dust3r_cvpr24}, while VGGT extends this paradigm to unordered multi-view image collections \cite{wang2025vggt}.
Later methods such as Depth Anything \cite{depthanything3}, $\pi^3$ \cite{wang2026pi}, and MapAnything \cite{keetha2026mapanything} improve the flexibility of this formulation.
However, these feed-forward visual geometry transformers often fail to scale to large image collections due to the quadratic complexity of their global attention mechanisms. Our work addresses this limitation by introducing compute-adaptive mixture-of-heads for visual geometry transformers that accelerates inference time while maintaining competitive reconstruction quality.

\paragraph{Accelerating Visual Geometry Transformers.}
Recent work has identified global attention in visual geometry transformers as a major source of computational overhead when processing long input sequences~\cite{shen2026fastvggt}.
Most existing methods resolve this bottleneck from a \emph{token redundancy} perspective, either by reducing the number of tokens or by sparsifying the key-value set used for global aggregation. 
For example, FastVGGT~\cite{shen2026fastvggt} merges redundant tokens before performing global attention, SparseVGGT~\cite{wang2025fastervggt} reduces the key-value set used for global aggregation, and TurboVGGT~\cite{huang2026turbovggt} compresses representative tokens for adaptive global attention.
These methods have recently been extended to operate at the head level. HTTM~\cite{wang2026httm} applies token merging independently within each head, while HeSS~\cite{Kim_2026_CVPR} performs head-wise key/value sparsification. 
However, both approaches remain focused on reducing tokens within individual heads and do not address architectural redundancy in global multi-head attention. 
Our work resolves the efficiency bottleneck from an \emph{architectural redundancy} perspective and proposes a compute-adaptive mixture-of-heads architecture that dynamically routes each attention head to different computation modes.
In addition, we find that our method can be combined with token reduction methods, such as token merging, for further speedups.

\paragraph{Mixture-of-Heads Transformers.}
A recent trend in efficient language models is using mixture-of-expert heads architecture that treat individual attention heads as conditional experts.
These methods are motivated by the observation that many attention heads contribute weakly to the final model output. 
Hence, they introduce routing networks that dynamically activate only a subset of heads for each input~\cite{zhang2022mixture,wu2024multihead,jin2024moh}.
Although mixture-of-heads methods are theoretically appealing, they typically reduce head importance to a binary keep-or-drop decision, overlooking intermediate-importance heads and the useful information retained even by low-saliency heads.
Our compute-adaptive mixture-of-heads architecture significantly differs from these methods in that it uses a lightweight router to adaptively estimate the compute required for each head and routes each head to different computation modes, each contributing to the final output representation of the attention block. 
This is designed for accelerating visual geometry transformers by reducing the architectural redundancy of global attention while maintaining competitive reconstruction quality.

\begin{figure*}[t!]
\centering
\includegraphics[width=0.99\textwidth]{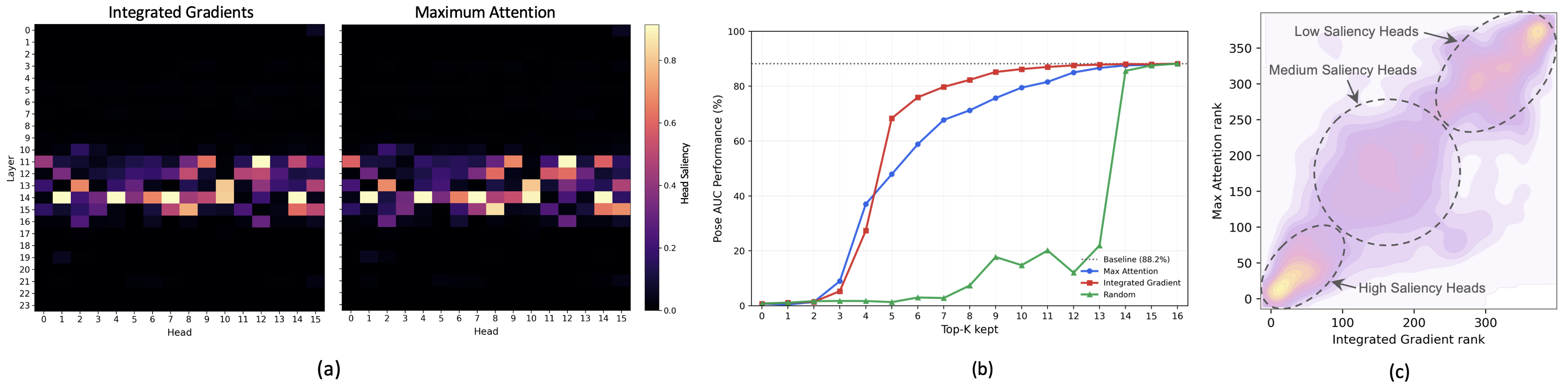}
\caption{{
    Head saliency analysis.
    (a) Head-saliency score matrix. Columns correspond to attention heads, while rows correspond to transformer layers. Brighter corresponds to more salient. 
    (b) Effect of pruning attention heads according to different saliency metrics on camera pose (AUC@30$^\circ$) on CO3Dv2.
    (c) Correlation between head rankings determined by maximum attention and integrated gradients. The rankings exhibit a Spearman's correlation coefficient of $\rho = 0.741$, $p < 0.001$.
    }}
\label{fig:architectural_redundancy_analysis}
\end{figure*}

\section{Architectural Redundancy Analysis}
\label{sec:architectural_redundancy_analysis}

We begin by briefly reviewing the components of visual geometry transformers that are most relevant to this work.
Unless otherwise specified, VGGT~\cite{wang2025vggt} serves as the primary backbone for both our analysis and proposed method.
VGGT consists of a DINOv2 encoder, a feature aggregator, and multiple task-specific prediction heads.
Among these components, the feature aggregator is the primary computational bottleneck \cite{shen2026fastvggt}.
Specifically, VGGT's feature aggregator comprises 24 global-attention layers, each with 16 attention heads, yielding a total of 384 global attention heads.
This raises a natural question: \textit{Are all these heads necessary for high-fidelity 3D reconstruction?}

\paragraph{Head Saliency.}
To investigate architectural redundancy in VGGT, we first analyze the saliency of individual attention heads for 3D reconstruction. 
We consider a head salient if at least one of its output tokens contributes substantially to the final scene representation. 
We study this using maximum integrated gradients (MIG) and maximum post-softmax attention (MaxAttn). 
MIG directly measures the influence of a head's output on the final representation, but remains a post-hoc gradient-based metric with no direct connection to the attention head architecture. 
MaxAttn instead reflects the head's query-key interactions and can therefore be predicted from its QKV features. 
Therefore, we use MIG to validate head importance and MaxAttn to guide our architecture. 
Although prior work estimates head saliency using Hessian-based metrics~\cite{Kim_2026_CVPR}, we find that our metrics provide a simpler and reliable saliency estimate.
Formal definitions and comparisons are provided in the appendix.

As shown in Figure~\ref{fig:architectural_redundancy_analysis}(a){\color{red}}, both metrics produce highly sparse and heterogeneous saliency maps.
This suggests that only a small subset of attention heads are important for 3D reconstruction.
We validate this observation in Figure~\ref{fig:architectural_redundancy_analysis}(b) by progressively pruning heads according to their saliency scores. Pruning a head means zeroing its output while retaining the residual pathway.
Removing attention heads in order of increasing saliency causes little degradation initially, followed by a sharp performance drop once highly salient heads are removed. Compared with random pruning, both metrics effectively identify salient heads, indicating that they capture a head saliency signal.

Lastly, we evaluate the agreement of the metrics through pairwise rank correlation. As shown in Figure~\ref{fig:architectural_redundancy_analysis}(c), the rankings achieve a Spearman correlation of $\rho = 0.741$, indicating strong consistency between the metrics. The plot also reveals three broad saliency clusters corresponding to high-saliency heads (ranks $\sim$1--100), medium-saliency heads ($\sim$101--250), and low-saliency heads ($\sim$251--384).
Together, the findings in Figure~\ref{fig:architectural_redundancy_analysis} reveal substantial redundancy among VGGT's global attention heads and hint at three saliency regimes, motivating our three-mode compute hierarchy.

\begin{figure*}[t!]
\centering
\includegraphics[width=0.99\textwidth]{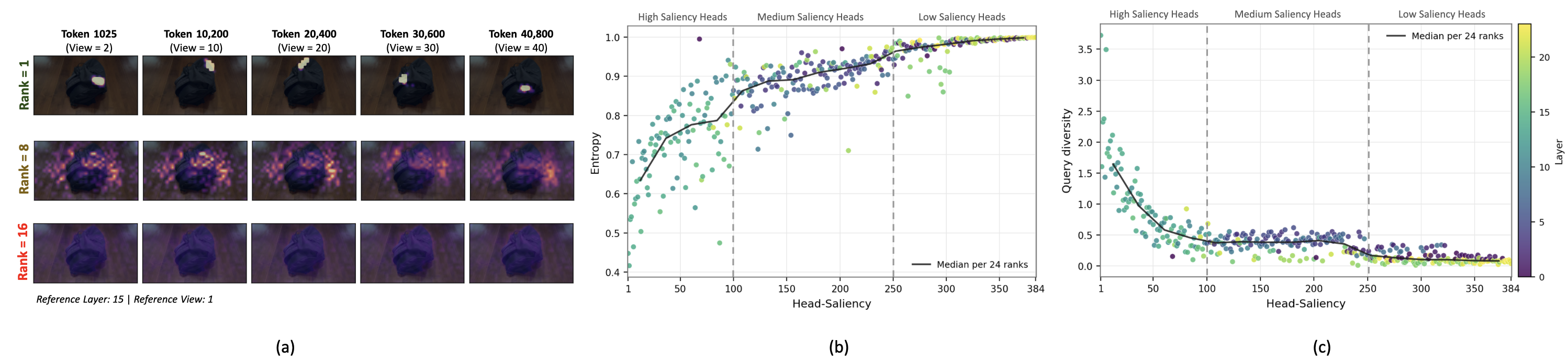}
\caption{{
    Head saliency analysis and attention-map visualizations.
    (a) Visualization of VGGT’s global attention maps from five query tokens (columns) to the tokens in the first view. Rank denotes each head’s saliency within layer 15.
    (b) Attention entropy as a function of head-saliency rank.
    (c) Query diversity as a function of head-saliency rank.
    Heads are ranked by maximum attention, ranging from most salient (rank 1) to least salient (rank 384).
    }}
\label{fig:architectural_redundancy_analysis_Quant}
\end{figure*}

\paragraph{Attention Patterns Across Saliency Regimes.}
To develop suitable replacement mechanisms for less salient heads, we next examine the attention patterns associated with each saliency regime. Figure~\ref{fig:architectural_redundancy_analysis_Quant}(a) provides a visualization of the attention maps from five query tokens towards the tokens in the first view.
The results show that low-saliency heads exhibit diffuse, nearly mean-pooling attention maps (bottom row). 
High-saliency heads, by contrast, produce focused, view-dependent attention patterns with substantially greater heterogeneity (top row). 
Medium-saliency heads attend to a small set of informative tokens (middle row), but these attention maps remain similar across query tokens.

{\paragraph{Quantifying Attention Concentration and Query Diversity.}
The qualitative observations in Figure~\ref{fig:architectural_redundancy_analysis_Quant}(a) suggest that head saliency is associated with two measurable properties: \emph{attention entropy} and \emph{query heterogeneity}. Attention entropy captures how concentrated a head's attention is over tokens, whereas query heterogeneity captures the variance of attention patterns across query tokens. We evaluate both properties against head saliency in Figure~\ref{fig:architectural_redundancy_analysis_Quant}(b) and (c).}

{For \emph{attention entropy}, we compute the Shannon entropy of each head's attention matrix, average it across inputs, and plot it against saliency rank. As shown in Figure~\ref{fig:architectural_redundancy_analysis_Quant}(b), entropy increases as saliency decreases. High-saliency heads exhibit the lowest entropy, medium-saliency heads exhibit intermediate entropy, and low-saliency heads exhibit the largest entropy (near 1). This confirms that low-saliency heads behave as mean pooling operators, suggesting that low-saliency heads can be approximated using mean pooling.}

{We next measure \emph{query diversity} using the mean channel-wise variance across query tokens. As shown in Figure~\ref{fig:architectural_redundancy_analysis_Quant}(c), high-saliency heads exhibit substantially greater query diversity than medium- and low-saliency heads. This indicates that salient heads produce strongly query-dependent attention maps, whereas less salient heads apply largely similar attention patterns across tokens. The reduced query diversity of medium-saliency heads suggests that their outputs can be approximated using a global query token.} 

{\paragraph{Summary.} Together, these results reveal a computational hierarchy. \emph{Low-saliency} heads have high entropy and low query diversity, indicating diffuse, token-invariant aggregation that can be replaced by mean pooling. \emph{Medium-saliency} heads retain concentrated attention but exhibit limited query diversity, suggesting that a single pooled query can capture their shared attention structure. \emph{High-saliency} heads combine low entropy with high query diversity and hence require full attention to preserve correspondences. }

\section{VGGT-Prime}
\label{sec:vggt-prime}

\begin{figure*}[t!]
\centering
\includegraphics[width=0.9\textwidth]{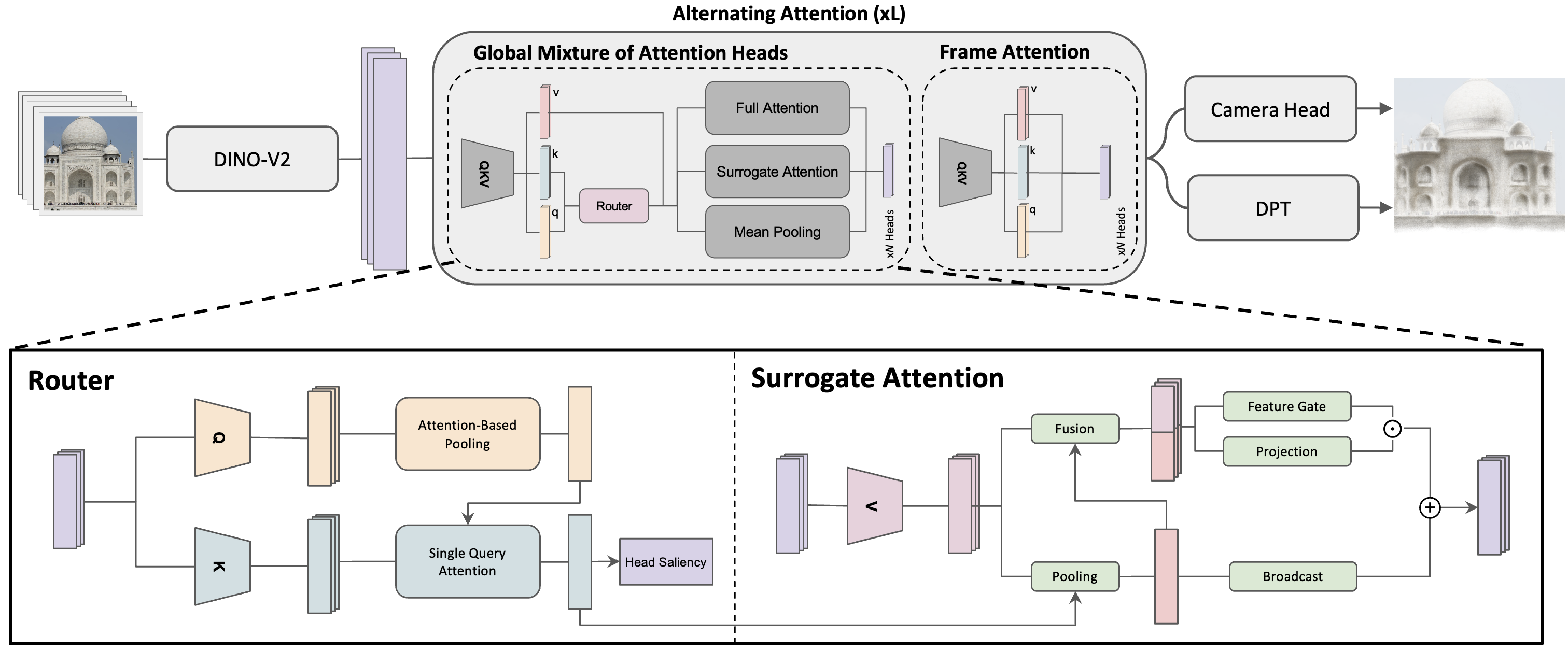}
\caption{An overview of VGGT-Prime's architecture.
VGGT-Prime uses a head-wise router to assign each global attention head to mean pooling, a learned surrogate attention, or exact softmax attention based on its estimated saliency.}
\label{fig1}
\end{figure*}

Motivated by our head-saliency analysis, we introduce \emph{VGGT-Prime}, an efficient mixture-of-heads architecture for visual  geometry transformers.
Figure~\ref{fig1} provides an overview of the proposed architecture.
Unlike standard mixture-of-head methods \cite{wu2024multihead}, which adopt a binary keep-or-drop strategy, VGGT-Prime adopts a head-wise router that estimates the global head-saliency and assigns each attention head to one of three computation modes: mean pooling, surrogate attention, or exact softmax attention.
This yields a compute-adaptive mixture of heads that replaces redundant heads with cheaper approximations while preserving full attention for the most salient heads.

\subsection{Head Saliency}
\label{sec:vggt-prime-head-saliency}

We begin by discussing VGGT-Prime's lightweight router used to estimate the saliency of each global attention head and assign it to an appropriate computation path.
Specifically, the router is built on top of our maximum-attention saliency metric and operates directly in QK space.
For each attention head ($h$), let ($Q_h, K_h, V_h \in \mathbb{R}^{N \times d}$) denote the query, key, and value feature matrices, respectively, where ($N$) is the number of input tokens and ($d$) is the internal dimension of head ($h$).
Rather than computing the full attention matrix, the router first compresses the query tokens into a single head-level query using an attention pooling module borrowed from attention-based multiple instance learning \cite{ITW:2018}.
For token \(i\) and head \(h\), we compute
\begin{equation}
\alpha_{h,i}
=
\operatorname{Softmax}(
w_h^\top
\left(
\tanh\!\left(W_h q_{h,i}\right)
\odot
\sigma\!\left(U_h q_{h,i})\right)
\right),
\end{equation}
where \(W_h\), \(U_h\), and \(w_h\) are learned pooling parameters, \(\sigma(\cdot)\) denotes the sigmoid function, and \(\odot\) denotes element-wise multiplication.
The pooled query is defined as:
\begin{equation}
q_h^{\mathrm{pool}}
=
\sum_{i=1}^{N} \alpha_{h,i} q_{h,i},
\qquad
\text{where}{\quad}q_h^{\mathrm{pool}} \in \mathbb{R}^{d}.
\end{equation}
We then use this pooled query to perform single-query attention over all keys, producing a routing vector $(r_h \in \mathbb{R}^{1 \times N})$ that captures the global attention pattern of head ($h$).
Finally, we define the head saliency score as the maximum routing value:
\begin{equation}
s_h = \max_{1 \leq i \leq N} r_h(i).
\end{equation}
In effect, this is similar to reducing the attention matrix along the row dimension to a single vector and then taking the maximum over the vector's entries. When trained to match the global attention distribution, the routing vector approximates the head-saliency ranking induced by full attention. We elaborate on training requirements in a later section.

\subsection{Head Routing}
\label{sec:vggt-prime-head-routing}

The saliency score \(s_h\) is used to route each head to one of three computation paths.
We use two fixed thresholds (\(\tau_{\mathrm{low}}=0.01\) and \(\tau_{\mathrm{high}}=0.20\)) which are optimized on a CO3Dv2 validation dataset and then held constant across all datasets.
Additional routing-threshold ablations and cross-dataset generalization experiments are provided in the appendix.
The routed head output is:
\begin{equation}
g_h(X)
=
\begin{cases}
\mathcal{A}_h(X),
& s_h \geq \tau_{\mathrm{high}}, \\[2mm]
\widetilde{\mathcal{A}}_h(X),
& \tau_{\mathrm{low}} < s_h < \tau_{\mathrm{high}}, \\[2mm]
\mathcal{M}_h(X),
& s_h \leq \tau_{\mathrm{low}}.
\end{cases}
\label{eq:vggt-prime-routing}
\end{equation}
Here, \(\mathcal{A}_h\) denotes the original VGGT softmax-attention operator for head \(h\), \(\widetilde{\mathcal{A}}_h\) denotes our learned surrogate attention, and \(\mathcal{M}_h\) denotes mean pooling over value tokens.

\subsection{Surrogate Attention for Medium-Saliency Heads}
\label{sec:vggt-prime-surrogate}

Medium-saliency heads exhibit attention patterns that are selective, yet relatively homogeneous across inputs.
We therefore approximate these heads using a single learned query attention pooling, followed by a token-dependent residual correction.
For each head \(h\), the router produces global routing weights \(r_h\), which are applied to the value tokens \(V_h\) to pool token information into a single query context vector:
\begin{equation}
C_h = r_h V_h,
\qquad
C_h \in \mathbb{R}^{1 \times d}.
\end{equation}
To account for low amplitude token-wise variation, we learn a residual correction term.
For each token \(i\), we concatenate its value feature and the pooled context $z_{h,i} = \left[ v_{h,i}, C_h\right]$.
A lightweight MLP predicts a correction term, which is modulated channel-wise via a gate:
\begin{equation}
\gamma_{h,i}
=
\sigma\!\left(\operatorname{MLP}_{\gamma}(z_{h,i})\right),
\qquad
\delta_{h,i}
=
\operatorname{MLP}_{\delta}(z_{h,i}).
\end{equation}
The final surrogate output for token \(i\) is
\begin{equation}
\widehat{y}_{h,i}
=
C_h 
+ 
\gamma_{h,i}
\odot
\delta_{h,i}.
\end{equation}

\subsection{Optimization of VGGT-Prime}

VGGT-Prime uses a two-stage distillation strategy. First, the surrogate and router are trained to match the latent features of a frozen VGGT teacher. Next, the surrogate and full-attention branches are jointly fine-tuned using soft routing. Finally, the mean-pooling branch is introduced without further training.

\paragraph{Router Optimization.}
The router is trained to capture the extrema structure of the attention matrix.
To achieve this, we max-pool the teacher's post-softmax attention matrix ($A_h \in \mathbb{R}^{N_k \times N_q}$) over queries:
\begin{equation}
p^{T}_{h,k}=\max_{q} A_{h,k,q}.
\end{equation}
This target identifies keys that receive strong attention from any key while preserving their relative scale. After normalization ($\bar r_{h}$ and $\bar p_{h}$), we define the router loss as
\begin{equation}
\mathcal{L}_{\mathrm{router}}^{(h)}
=
D_{\mathrm{KL}}\!\!\left(\bar p_{h}^{T}
\,\middle\|\,\bar  r_{h}\right) +
\lambda \left| \max_k p_{h,k}^{T} - \max_k r_{h,k} \right|.
\end{equation}

\paragraph{Surrogate Attention Optimization.}

To distill the teacher (T) latents into the surrogate attention, we train the surrogate at each head ($h$) to predict the residual update:
\begin{equation}
\Delta_{h}^{\mathrm{T}}=(A_{h}-I)V_{h}.
\end{equation}
This delta formulation supervises the non-identity component of attention, while decoupling the training objective across layers.
We define the latent matching loss as: 

\begin{equation}
\mathcal{L}_{\mathrm{latent}}^{(h)}
=
\left\|
\widehat{\Delta}_{h}
-
\Delta_{h}^{\mathrm{T}}
\right\|_{2}
+
\lambda
\left(1-\operatorname{cos}\!\left(
\widehat{\Delta}_{h},
\Delta_{h}^{\mathrm{T}}
\right)\right).
\end{equation}

\section{Experiments}

We evaluate VGGT-Prime on point-cloud reconstruction using summed Chamfer distance (CD), camera-pose estimation using Relative Translation Accuracy (RTA) and AUC at $30^\circ$, and depth estimation using absolute relative error (AbsRel) and threshold accuracy ($\delta < 1.25$). These metrics are standard in prior work~\cite{shen2026fastvggt,wang2025fastervggt}. Additional experimental details are provided in the appendix.

\begin{table*}[t!]
\centering
\small
\setlength{\tabcolsep}{4pt}
\renewcommand{\arraystretch}{1.12}
\resizebox{0.99\textwidth}{!}{
\begin{tabular}{l|cc|cc|cc|cc|cc}
\toprule
\multirow{2}{*}{\textbf{Method}}
& \multicolumn{2}{c|}{\textbf{7-Scenes Sparse}}
& \multicolumn{2}{c|}{\textbf{7-Scenes Dense}}
& \multicolumn{2}{c|}{\textbf{ScanNet 100}}
& \multicolumn{2}{c|}{\textbf{ScanNet 500}}
& \multicolumn{2}{c}{\textbf{ETH3D}} \\
\cmidrule(lr){2-3}
\cmidrule(lr){4-5}
\cmidrule(lr){6-7}
\cmidrule(lr){8-9}
\cmidrule(lr){10-11}
&
\textbf{CD $\downarrow$}
& \textbf{Time (s) $\downarrow$}
&
\textbf{CD $\downarrow$}
& \textbf{Time (s) $\downarrow$}
&
\textbf{CD $\downarrow$}
& \textbf{Time (s) $\downarrow$}
&
\textbf{CD $\downarrow$}
& \textbf{Time (s) $\downarrow$}
&
\textbf{CD $\downarrow$}
& \textbf{Time (s) $\downarrow$} \\
\midrule

VGGT (CVPR 2025)
& 0.118 & 4.5
& 0.115 & 38.1
& 0.421 & 4.9
& 0.442 & 90.1
& 1.07 & \underline{0.39} \\

\midrule

FastVGGT (ICLR 2026)
& 0.109 & \underline{\underline{2.7}}
& 0.115 & \underline{\underline{14.2}}
& 0.416 & \underline{\underline{2.8}}
& 0.437 & \underline{\underline{28.4}}
& 1.24 & \underline{\underline{0.42}} \\

SparseVGGT (CVPR 2026)
& 0.134 & \underline{2.6}
& 0.139 & 16.2
& 0.425 & \underline{2.6}
& 0.441 & 34.2
& 1.89 & 0.51 \\

HTTM (CVPR 2026)
& 0.121 & 3.8
& 0.117 & \underline{12.1}
& 0.420 & 4.0
& 0.433 & \underline{22.8}
& 1.41 & 0.73 \\

HeSS (CVPR 2026)
& 0.126 & 2.8
& 0.168 & 18.4
& 0.418 & 2.9
& 0.435 & 37.4
& 1.67 & 0.57 \\

\midrule

VGGT-Prime (Ours)
& 0.121 & \textbf{1.9}
& 0.115 & \textbf{9.8}
& 0.422 & \textbf{2.0}
& 0.441 & \textbf{19.2}
& 0.99 & \textbf{0.38} \\

\bottomrule
\end{tabular}
}
\caption{
Point-cloud reconstruction comparison on ScanNet, 7-Scenes, and ETH3D. Bold, \underline{underlined}, and \underline{\underline{double-underlined}} values indicate the fastest, second-fastest, and third-fastest models, respectively. HTTM and HeSS are PyTorch re-implementations.
}
\label{tab:pointcloud_all_results}
\end{table*}

\begin{table}[t!]
\centering
\small
\setlength{\tabcolsep}{1.5pt}
\renewcommand{\arraystretch}{1.12}

\begin{tabular}{l|ccc|ccc}
\toprule
\multirow{2}{*}{\textbf{Method}}
& \multicolumn{3}{c|}{\textbf{7-Scenes}}
& \multicolumn{3}{c}{\textbf{CO3Dv2}} \\
\cmidrule(lr){2-4}
\cmidrule(lr){5-7}
&
\textbf{RTA $\uparrow$}
& \textbf{AUC $\uparrow$}
& \textbf{Time $\downarrow$}
&
\textbf{RTA $\uparrow$}
& \textbf{AUC $\uparrow$}
& \textbf{Time $\downarrow$} \\
\midrule

VGGT
& 96.7 & 79.8 & 38.1
& 97.3 & 90.4 & 4.8 \\

\midrule

FastVGGT
& 96.4 & 79.2
& \underline{\underline{14.2}}
& 99.0 & 84.1
& \underline{\underline{2.8}} \\

SparseVGGT
& 95.3 & 74.2
& 16.2
& 98.9 & 83.8
& \underline{2.7} \\

HTTM
& 94.3 & 75.4 & \underline{12.1}
& 96.8 & 88.1 & 4.3 \\

HeSS
& 93.9 & 73.9 & 18.4
& 96.9 & 86.5 & 3.4 \\

\midrule

VGGT-Prime
& 95.1 & 78.6
& \textbf{9.8}
& 99.2 & 87.2
& \textbf{2.0} \\

\bottomrule
\end{tabular}

\caption{
Pose estimation quality and runtime comparison.
}
\label{tab:pose_quality_single}
\end{table}
\begin{table}[t!]
\centering
\small
\setlength{\tabcolsep}{1mm}
\renewcommand{\arraystretch}{1.12}

\begin{tabular}{l|ccc|cc}
\toprule
\multirow{2}{*}{\textbf{Method}}
& \multicolumn{3}{c|}{\textbf{7-Scenes}}
& \multicolumn{2}{c}{\textbf{Sintel}} \\
\cmidrule(lr){2-4}
\cmidrule(lr){5-6}
&
\textbf{AbsRel $\downarrow$}
& \textbf{$\delta$ $\uparrow$}
& \textbf{Time(s)}
&
\textbf{AbsRel $\downarrow$}
& \textbf{$\delta$ $\uparrow$} \\
\midrule

VGGT
& 0.271 & 0.958 & 38.1
& 0.592 & 0.583 \\

\midrule

FastVGGT
& 0.272 & 0.960
& \underline{\underline{14.2}}
& 0.620 & 0.526 \\

SparseVGGT
& 0.274 & 0.962
& 16.2
& 0.675 & 0.518 \\

HTTM
& 0.276 & 0.959
& \underline{12.1}
& 0.647 & 0.538 \\

HeSS
& 0.277 & 0.962 & 18.4
& 0.577 & 0.520 \\

\midrule

VGGT-Prime
& 0.275 & 0.965
& \textbf{9.8}
& 0.524 & 0.591 \\

\bottomrule
\end{tabular}

\caption{
Depth estimation quality and runtime comparison.
}
\label{tab:depth_quality_single}
\end{table}

\subsection{Comparison with State-of-the-Art Methods}
We compare VGGT-Prime against VGGT and state-of-the-art efficient 3D foundation models. For FastVGGT~\cite{shen2026fastvggt} and SparseVGGT~\cite{wang2025fastervggt}, we use the publicly available implementations and checkpoints. We additionally reimplement HTTM~\cite{wang2026httm} and HeSS~\cite{Kim_2026_CVPR} in PyTorch. For brevity, we report results on five of the seven benchmark datasets in the main text, with complete details provided in the appendix.

\paragraph{Point Cloud Reconstruction.}
We first evaluate VGGT-Prime on point-cloud reconstruction to assess its overall geometric fidelity. Table~\ref{tab:pointcloud_all_results} reports results on ScanNet~\cite{dai2017scannet}, 7-Scenes~\cite{shotton2013scene}, and ETH3D~\cite{Schops_2019_CVPR}, which span diverse evaluation settings, including large- and small-scale environments and complex real-world imagery.
Across all datasets, VGGT-Prime achieves the lowest runtime while maintaining competitive reconstruction quality. On ScanNet, it is up to $4.7\times$ faster than VGGT and $1.2\times$ faster than the fastest efficient baseline, HTTM. On dense 7-Scenes with stride (3), VGGT-Prime achieves a $3.9\times$ speedup over VGGT while matching its reconstruction quality. Under sparse sampling with stride (10), it remains the fastest method, delivering a $2.4\times$ speedup over VGGT while preserving competitive reconstruction quality. On ETH3D with 20 views, VGGT-Prime achieves both the best reconstruction performance and the lowest runtime.
Overall, these results show that VGGT-Prime substantially reduces VGGT’s inference time while closely preserving its reconstruction quality.

{\paragraph{Camera Pose Estimation.}
Next, we evaluate camera pose estimation, which provides a strong test of whether the model preserves cross-view correspondences in global attention \cite{wang2026vggtomega}. Table~\ref{tab:pose_quality_single} highlights pose estimation performance on dense 7-Scenes and dense CO3Dv2~\cite{reizenstein2021common}. 
On 7-Scenes, VGGT-Prime achieves a {\(3.9\times\)} speedup over VGGT with only minor reductions in RTA and AUC. 
On CO3Dv2, it achieves the highest RTA and an AUC competitive with VGGT while being {\(2.4\times\)} faster.
Overall, these results demonstrate that VGGT-Prime preserves the global correspondence structure required for pose estimation, while being the fastest model.}

{\paragraph{Depth Estimation.}
Lastly, we evaluate depth estimation, which relies on global attention to refine per-view predictions using global scene context \cite{wang2026vggtomega}. 
Table~\ref{tab:depth_quality_single} evaluates depth estimation on dense 7-Scenes and pairwise Sintel~\cite{butler2012sintel}.
On 7-Scenes, VGGT-Prime achieves the highest \(\delta\) accuracy while being {\(3.9\times\)} faster than VGGT, with only a minor increase in AbsRel. On Sintel, VGGT-Prime achieves the best performance under both metrics. Overall, these results indicate that VGGT-Prime preserves depth prediction quality while reducing runtime.}


\begin{table}[t!]
\centering
\small
\setlength{\tabcolsep}{5pt}
\renewcommand{\arraystretch}{1.15}

\begin{tabular}{lcccc}
\toprule
\textbf{Variant}
& \textbf{CD} $\downarrow$
& \textbf{AUC@30} $\uparrow$
& \textbf{AbsRel} $\downarrow$
& \textbf{Time (s)} \\
\midrule
VGGT-$\Omega$
& 0.10
& 88.2
& 0.275
& 2.9 \\
VGGT-$\Omega$-Prime
& 0.10
& 86.3
& 0.284
& \textbf{1.6} \\
\midrule
$\pi^3$
& 0.11
& 82.4
& 0.327
& 3.3 \\
$\pi^3$-Prime
& 0.11
& 81.9
& 0.362
& \textbf{1.5} \\

\midrule
DA3
& 0.10
& 82.4
& 0.305
& 3.5 \\
DA3-Prime
& 0.10
& 79.3
& 0.320
& \textbf{1.5} \\
\bottomrule
\end{tabular}
\caption{
Generalization across backbones on 7-scenes.
}
\label{tab:generalization_to_other_transformers}
\end{table}

\subsection{Generalization to Other Backbones}
We apply Prime to the global-attention blocks of three additional visual geometry transformers on sparse 7-Scenes, including VGGT-$\Omega$~\cite{wang2026vggtomega}, $\pi^3$~\cite{wang2026pi}, and Depth Anything 3 (DA3)~\cite{depthanything3}. For VGGT-$\Omega$, we distill the non-register global-attention blocks; for $\pi^3$, the cross-view blocks in the odd-indexed decoder layers; and for DA3, the global-attention blocks in the image encoder. As shown in Table~\ref{tab:generalization_to_other_transformers}, Prime provides approximately $2\times$ speedup across all backbones on sparse 7-scenes. These results highlight the generalizability of our Prime approach.

\subsection{Ablation Studies}

\paragraph{Component Ablation.}
To assess the relative contribution of the components in VGGT-Prime's architecture, we conduct a component ablation. 
Specifically, we replace the router’s attention-based query pooling (Q-Pooling) with mean pooling, replace the surrogate context vector (Context) with a mean-pooled value vector, and remove the token-wise residual correction module (Residual).
As shown in Table~\ref{tab:component_ablation}, removing the residual correction causes a performance degradation of 44.6\%. Replacing the context-vector formulation results in a 33.1\% drop, while replacing attention-based query pooling produces a degradation of 14.6\%.
This suggests that residual correction is particularly important at high pruning ratios, where some salient heads may be replaced by surrogate attention and their token-dependent variability must be preserved. 
Overall, these results show the importance of all three components.


\begin{table}[t!]
\centering
\small
\setlength{\tabcolsep}{5pt}
\renewcommand{\arraystretch}{1.15}

\begin{tabular}{lccc|c}
\toprule
\textbf{Variant}
& \textbf{Context}
& \textbf{Q-Pooling}
& \textbf{Residual}
& \textbf{AUC@30} \\
\midrule
Base
& \checkmark
& \checkmark
& \checkmark
& \textbf{87.2} \\
\midrule
Mean context
& \xmark
& \checkmark
& \checkmark
& 54.1 \\
Mean query
& \checkmark
& \xmark
& \checkmark
& 72.9 \\
No residual
& \checkmark
& \checkmark
& \xmark
& 42.6 \\
\bottomrule
\end{tabular}
\caption{
Component ablation of VGGT-Prime on CO3Dv2.
}
\label{tab:component_ablation}

\end{table}

\begin{table}[t!]
\centering
\small
\setlength{\tabcolsep}{5pt}
\renewcommand{\arraystretch}{1.15}

\begin{tabular}{lccccc}
\toprule
\textbf{Variant}
& \textbf{CD} $\downarrow$
& \textbf{AUC@30} $\uparrow$
& \textbf{AbsRel} $\downarrow$
& \textbf{Time (s)} \\
\midrule
Base
& \textbf{0.115}
& \textbf{78.6}
& \textbf{0.275}
& 9.8 \\
\midrule
Binary Routing
& 0.581
& 40.9
& 0.452
& \textbf{6.6} \\
Linear Attention
& 0.153
& 73.7
& 0.274
& 11.9 \\
\bottomrule
\end{tabular}
\caption{
Linear-attention ablation on dense 7-Scenes.
}
\label{tab:linear_attention_ablation}

\end{table}

\paragraph{Compute Mode Ablation.}
We examine whether the proposed three-branch routing scheme can be replaced by simpler alternatives. Specifically, we compare VGGT-Prime against a variant that replaces the surrogate attention with linear attention~\cite{katharopoulos2020transformers} and a two-branch variant that removes the surrogate attention, retaining only mean pooling and full attention. As shown in Table~\ref{tab:linear_attention_ablation}, both alternatives underperform VGGT-Prime, indicating that the learned surrogate is better suited to capturing VGGT's sparse, head-specific correspondences.

\section{Conclusion}
In this paper, we present VGGT-Prime, a compute-adaptive mixture-of-heads model for efficient feed-forward 3D reconstruction.
We empirically reveal the \emph{architectural redundancy} in the global multi-head attention mechanism of visual geometry transformers 
and present a method to resolve this redundancy by \emph{adaptively routing} each attention head to an appropriate computation mode to reduce computational cost while maintaining reconstruction quality.
Extensive experiments demonstrate the superiority of VGGT-Prime over state-of-the-art methods.

{
\bibliographystyle{plainnat}
\bibliography{ref}
}

\section*{Technical Appendix}
\addcontentsline{toc}{section}{Appendix}

\section{Limitation and Future Work}
\label{app:limitations}
One limitation of our work is that it relies on a frozen teacher model to provide distillation targets for training the router and surrogate attention.
This is effective for mimicking the teacher model, but it also limits the method to matching the teacher rather than surpassing it.
Our future work aims to explore architectural modifications that allow for training directly on ground truth.

\section{Implementation Details}
\label{app:Implementation}

\paragraph{Training Dataset.}
We train VGGT-Prime on a mixture of real, synthetic, indoor, and outdoor multi-view datasets, including BlendedMVS \cite{yao2020blendedmvs}, Mapillary Planet-Scale Depth \cite{lopezantequera2020mapillary}, ScanNet++ v2 \cite{yeshwanth2023scannetpp}, Spring \cite{mehl2023spring}, TartanAirV2-WB \cite{wang2020tartanair}, UnrealStereo4K \cite{tosi2021smdnets}, Aria Synthetic Environments \cite{aria_synthetic_environments}, DL3DV-10K \cite{ling2024dl3dv10k}, Dynamic Replica \cite{karaev2023dynamicstereo}, MegaDepth \cite{li2018megadepth}, MVS-Synth \cite{huang2018deepmvs}, ParallelDomain-4D \cite{paralleldomain4d}, and SAIL-VOS 3D \cite{hu2021sailvos3d}.

\paragraph{Hyperparameters.}
All hidden dimensions in the surrogate attention and router are set to $64$. Both modules are trained for $10$ epochs using AdamW with an initial learning rate of $10^{-3}$, weight decay of $10^{-4}$, dropout of $0.1$, and a cosine-annealing learning schedule. All seeds are set to $67$. We use routing thresholds $\tau_{\mathrm{low}}=0.01$ and $\tau_{\mathrm{high}}=0.2$.

\paragraph{Runtime measurement.}
Runtime is reported as synchronized model inference time in seconds. 
For a fair comparison, timing includes the encoder, aggregator, and prediction heads. 
We report the median runtime after three warm-up iterations. 
All methods are evaluated using the same hardware, input resolution, BF16 percision, and a batch size of 1. 
Unless otherwise stated, all baselines use their publicly available implementations.
For VGGT, we use the accelerated implementation introduced by \citet{keetha2026mapanything}.

\paragraph{Packages.} Where possible, we used enviornments with Python v3.12.13. Data aggregation and metric computation used NumPy (v2.4.4), Pandas (v3.0.3), SciPy (v1.17.1), and Scikit-Learn (v1.8.0). Model inference and profiling used PyTorch (v2.12.0+cu130; CUDA v13.0 build), TorchVision (v0.27.0), Triton (v3.7.0), xFormers (v0.0.35), and timm (v1.0.27). Image loading and preprocessing used OpenCV (v4.13.0.92), Pillow (v12.2.0), scikit-image (v0.26.0), and imageio (v2.37.3). 3D reconstruction and point-cloud evaluation used Open3D (v0.19.0), PyCOLMAP (v3.10.0), Trimesh (v4.12.2), and evo (v1.36.4). 

\paragraph{VGGT-Prime Algorithm.}
To clarify VGGT-Prime's implementation, we provide the pseudo-code for its global aggregator. All other components are borrowed from VGGT.

\begin{algorithm}[h!]
\caption{VGGT-Prime: Global Aggregator Block}
\label{alg:vggt_prime}

\textbf{Input:} Tokens \(X^{(\ell)} \in \mathbb{R}^{N \times D}\), positional information \(P\), and thresholds
\(\tau_{\mathrm{low}}, \tau_{\mathrm{high}}\). \\

\textbf{Output:} Updated tokens \(X^{(\ell+1)}\).

\vspace{0.5em}
\textbf{(1) Normalize input tokens}

\[
\widetilde{X}^{(\ell)}
\leftarrow
\operatorname{LN}_1^{(\ell)}
\left(
X^{(\ell)}
\right).
\]

\vspace{0.5em}
\textbf{(2) Compute routed attention}

\[
\begin{aligned}
(Q_h,K_h,V_h)
&\leftarrow
\operatorname{QKV}_h^{(\ell)}
\left(
\widetilde{X}^{(\ell)}, P
\right),\\
s_h^{(\ell)}
&\leftarrow
f_{\mathrm{route}}^{(\ell)}
\left(
Q_h,K_h
\right),
\end{aligned}
\]

\[
Y_h =
\begin{cases}
\displaystyle
\operatorname{softmax}\!\left(
\frac{Q_h K_h^\top}{\sqrt{d}}
\right)V_h,
&
s_h^{(\ell)} \geq \tau_{\mathrm{high}},
\\[3mm]
\widetilde{f}_h^{(\ell)}(Q_h,K_h,V_h),
&
\tau_{\mathrm{low}}
<
s_h^{(\ell)}
<
\tau_{\mathrm{high}},
\\[3mm]
\displaystyle
\operatorname{Broadcast}\!\left(
\frac{1}{N}
\sum_{i=1}^{N} V_{h,i}
\right),
&
s_h^{(\ell)} \leq \tau_{\mathrm{low}} .
\end{cases}
\]

\vspace{0.5em}
\textbf{(3) Merge routed heads and apply attention residual}

\[
Y
\leftarrow
\operatorname{Concat}(Y_1,\dots,Y_H),
\qquad
A^{(\ell)}
\leftarrow
\operatorname{LS}_1^{(\ell)}
\left(
Y W_O^{(\ell)}
\right).
\]

\[
X_{\mathrm{attn}}^{(\ell)}
\leftarrow
X^{(\ell)} + A^{(\ell)} .
\]

\vspace{0.5em}
\textbf{(4) Apply feed-forward block}

\[
\widetilde{Z}^{(\ell)}
\leftarrow
\operatorname{LN}_2^{(\ell)}
\left(
X_{\mathrm{attn}}^{(\ell)}
\right).
\]

\[
F^{(\ell)}
\leftarrow
\operatorname{LS}_2^{(\ell)}
\left(
\operatorname{FFN}^{(\ell)}
\left(
\widetilde{Z}^{(\ell)}
\right)
\right).
\]

\[
X^{(\ell+1)}
\leftarrow
X_{\mathrm{attn}}^{(\ell)} + F^{(\ell)} .
\]

\vspace{0.5em}
\textbf{return} \(X^{(\ell+1)}\)
\end{algorithm} 
\begin{table*}[t!]
\centering
\small
\setlength{\tabcolsep}{2.5pt}
\renewcommand{\arraystretch}{1.12}
\resizebox{0.99\textwidth}{!}{
\begin{tabular}{l|cccc|cccc|cccc}
\toprule
\multirow{2}{*}{\textbf{Method}}
& \multicolumn{4}{c|}{\textbf{ScanNet 100}}
& \multicolumn{4}{c|}{\textbf{ScanNet 300}}
& \multicolumn{4}{c}{\textbf{ScanNet 500}} \\
\cmidrule(lr){2-5}
\cmidrule(lr){6-9}
\cmidrule(lr){10-13}
&
\textbf{Acc. $\downarrow$}
& \textbf{Comp. $\downarrow$}
& \textbf{CD $\downarrow$}
& \textbf{Time $\downarrow$}
&
\textbf{Acc. $\downarrow$}
& \textbf{Comp. $\downarrow$}
& \textbf{CD $\downarrow$}
& \textbf{Time $\downarrow$}
&
\textbf{Acc. $\downarrow$}
& \textbf{Comp. $\downarrow$}
& \textbf{CD $\downarrow$}
& \textbf{Time $\downarrow$} \\
\midrule

VGGT (CVPR 2025)
& 0.198 & 0.223 & 0.421 & 4.9
& 0.200 & 0.229 & 0.429 & 33.6
& 0.205 & 0.236 & 0.442 & 90.1 \\

\midrule

Fast3R (CVPR 2025)
& 0.247 & 0.288 & 0.535 & {\underline{2.5}}
& 0.256 & 0.297 & 0.551 & 17.2
& 0.263 & 0.297 & 0.560 & 62.5 \\

TTT3R (ICLR 2026)
& 0.218 & 0.236 & 0.455 & 4.3
& 0.228 & 0.250 & 0.477 & \underline{\underline{12.1}}
& 0.239 & 0.258 & 0.497 & \underline{19.8} \\

HTTM (CVPR 2026)
& 0.195 & 0.225 & 0.420 & 4.0
& 0.202 & 0.226 & 0.428 & \underline{11.7}
& 0.204 & 0.229 & 0.433 & \underline{\underline{22.8}} \\

HeSS (CVPR 2026)
& 0.199 & 0.219 & 0.418 & 2.9
& 0.205 & 0.224 & 0.429 & 15.5
& 0.207 & 0.228 & 0.435 & 37.4 \\

FastVGGT (ICLR 2026)
& 0.197 & 0.219 & 0.416 & {2.8}
& 0.201 & 0.227 & 0.430 & 12.3
& 0.210 & 0.217 & 0.437 & 28.4 \\

SparseVGGT (CVPR 2026)
& 0.202 & 0.222 & 0.425 & \underline{\underline{2.6}}
& 0.207 & 0.231 & 0.438 & 14.2
& 0.209 & 0.232 & 0.441 & 34.2 \\

\midrule

VGGT-Prime (Ours)
& 0.194 & 0.228 & 0.422 & \textbf{2.0}
& 0.207 & 0.225 & 0.432 & \textbf{8.6}
& 0.220 & 0.221 & 0.441 & \textbf{19.2} \\

\bottomrule
\end{tabular}
}

\caption{
    Point cloud reconstruction performance on ScanNet with 100, 300, and 500 input views.
}
\label{tab:scannet_appendix}

\end{table*}
\begin{table*}[h!]
\centering
\small
\setlength{\tabcolsep}{2.5pt}
\renewcommand{\arraystretch}{1.12}
\resizebox{0.99\textwidth}{!}{
\begin{tabular}{l|cccc|cccc|cccc}
\toprule
\multirow{2}{*}{\textbf{Method}}
& \multicolumn{4}{c|}{\textbf{7-Scenes Sparse}}
& \multicolumn{4}{c|}{\textbf{7-Scenes Dense}}
& \multicolumn{4}{c}{\textbf{ETH3D}} \\
\cmidrule(lr){2-5}
\cmidrule(lr){6-9}
\cmidrule(lr){10-13}
&
\textbf{Acc. $\downarrow$}
& \textbf{Comp. $\downarrow$}
& \textbf{CD $\downarrow$}
& \textbf{Time $\downarrow$}
&
\textbf{Acc. $\downarrow$}
& \textbf{Comp. $\downarrow$}
& \textbf{CD $\downarrow$}
& \textbf{Time $\downarrow$}
&
\textbf{Acc. $\downarrow$}
& \textbf{Comp. $\downarrow$}
& \textbf{CD $\downarrow$}
& \textbf{Time $\downarrow$} \\
\midrule

VGGT (CVPR 2025)
& 0.056 & 0.062 & 0.118 & 4.5
& 0.053 & 0.061 & 0.115 & 38.1
& 0.504 & 0.566 & 1.07 & \underline{0.39} \\

\midrule

Fast3R (CVPR 2025)
& 0.095 & 0.118 & 0.213 & {\underline{2.3}}
& 0.102 & 0.121 & 0.223 & 21.2
& 1.195 & 1.217 & 2.41 & 1.02 \\

TTT3R (ICLR 2026)
& 0.072 & 0.091 & 0.163 & 4.5
& 0.066 & 0.053 & 0.119 & \underline{11.3}
& 1.021 & 0.963 & 1.98 & 1.40 \\

HTTM (CVPR 2026)
& 0.057 & 0.065 & 0.121 & 3.8
& 0.052 & 0.065 & 0.117 & \underline{\underline{12.1}}
& 0.671 & 0.736 & 1.41 & 0.73 \\

HeSS (CVPR 2026)
& 0.057 & 0.069 & 0.126 & 2.8
& 0.094 & 0.074 & 0.168 & 18.4
& 0.812 & 0.853 & 1.67 & 0.57 \\

FastVGGT (ICLR 2026)
& 0.051 & 0.059 & 0.109 & {2.7}
& 0.049 & 0.066 & 0.115 & 14.2
& 0.582 & 0.663 & 1.24 & \underline{\underline{0.42}} \\

SparseVGGT (CVPR 2026)
& 0.061 & 0.072 & 0.134 &  \underline{\underline{2.6}}
& 0.071 & 0.068 & 0.139 & 16.2
& 1.097 & 0.792 & 1.89 & 0.51\\

\midrule

VGGT-Prime (Ours)
& 0.066 & 0.054 & 0.121 & \textbf{1.9}
& 0.053 & 0.062 & 0.115 & \textbf{9.8}
& 0.506 & 0.483 & 0.99 & \textbf{0.38} \\

\bottomrule
\end{tabular}
}

\caption{
Point cloud reconstruction comparison on sparse and dense 7-Scenes and ETH3D.
}
\label{tab:other_pc}
\end{table*}
\section{Additional Comparisons Against SoTA}
\label{app:Benchmarks}

We extend the benchmark of VGGT-Prime by adding two more datasets, two more state-of-the-art (SoTA) models, and more detailed evaluation metrics. 
Specifically, we add RealEstate10K~\cite{reizenstein2021common} for pose estimation and Bonn~\cite{palazzolo2019refusion} for depth estimation, and include Fast3R~\cite{Yang_2025_Fast3R} and TTT3R~\cite{chen2026tttr} as additional SoTA models. 
We also report point-cloud accuracy (acc) and completeness (comp), along with relative rotation accuracy (RRA) for camera pose estimation.

\paragraph{Additional Baseline Implementation Details.} In this benchmark, we compare VGGT-Prime with FastVGGT, SparseVGGT, Fast3R, TTT3R, HTTM, and HeSS. 
For FastVGGT, SparseVGGT, Fast3R, and TTT3R, we use the publicly available implementations and checkpoints. 
However, since the official implementations of HTTM and HeSS are unavailable, we reimplement both methods in PyTorch to the best of our ability and validate our implementations by reproducing the reconstruction quality reported in the original papers~\cite{wang2026httm,Kim_2026_CVPR}.
Exact reproduction is not possible because several implementation details are underspecified. In particular, HTTM relies on a custom CUDA kernel whose implementation is not described, while HeSS does not fully specify its calibration and compute-budget allocation procedures. We therefore report the reconstruction quality and runtime of our PyTorch reimplementations while acknowledging these limitations.

\paragraph{Point Cloud Reconstruction Results.}

Across all datasets and view counts, VGGT-Prime is consistently faster than VGGT while maintaining comparable reconstruction quality. On ScanNet (Table~\ref{tab:scannet_appendix}), it achieves the best completeness at 100 frames and the best accuracy at 300 and 500 frames. Similar trends hold on 7-Scenes (Table~\ref{tab:other_pc}), while on ETH3D, VGGT-Prime achieves both the best overall reconstruction quality and the lowest runtime. More broadly, the competing efficient spatial foundation models generally preserve VGGT-level reconstruction quality but are slower than VGGT-Prime. 

Among the efficient SoTA models, TTT3R is the fastest, achieving a $4\times$ speedup over VGGT on ScanNet-500, followed closely by HTTM and HeSS. In contrast, Fast3R is substantially slower and exhibits a notably higher Chamfer distance. Among the head-wise methods, HTTM provides the strongest reconstruction quality, followed by HeSS. Moreover, at lower view counts, such as ETH3D, all newly added baselines are slower than VGGT, indicating considerable fixed computational overhead. Overall, VGGT-Prime provides the strongest efficiency-accuracy trade-off, substantially reducing runtime relative to both the original and newly added SoTA models.

\begin{table*}[h!]
\centering
\small
\setlength{\tabcolsep}{1.5pt}
\renewcommand{\arraystretch}{1.12}

\begin{adjustbox}{max width=\linewidth}
\begin{tabular}{l|cccc|cccc|ccc}
\toprule
\multirow{2}{*}{\textbf{Method}}
& \multicolumn{4}{c|}{\textbf{7-Scenes}}
& \multicolumn{4}{c|}{\textbf{CO3Dv2}}
& \multicolumn{3}{c}{\textbf{RE10K}} \\
\cmidrule(lr){2-5}
\cmidrule(lr){6-9}
\cmidrule(lr){10-12}
&
\textbf{RRA $\uparrow$}
& \textbf{RTA $\uparrow$}
& \textbf{AUC $\uparrow$}
& \textbf{Time $\downarrow$}
&
\textbf{RRA $\uparrow$}
& \textbf{RTA $\uparrow$}
& \textbf{AUC $\uparrow$}
& \textbf{Time $\downarrow$}
&
\textbf{RRA $\uparrow$}
& \textbf{RTA $\uparrow$}
& \textbf{AUC $\uparrow$} \\
\midrule

VGGT (CVPR 2025)
& 100.0 & 96.7 & 79.8 & 38.1
& 99.0 & 97.3 & 90.4 & 4.8
& 100.0 & 93.5 & 75.3 \\

\midrule

Fast3R (CVPR 2025)
& 90.0 & 80.4 & 56.5 & 21.2
& 96.7 & 93.4 & 78.4 & 4.7
& 100.0 & 88.0 & 63.3 \\

TTT3R (ICLR 2026)
& 99.9 & 91.7 & 62.9 & \underline{11.3}
& 96.0 & 91.9 & 73.0 & 4.6
& 100.0 & 96.0 & 77.3 \\

HTTM (CVPR 2026)
& 98.7 & 94.3 & 75.4 & \underline{\underline{12.1}}
& 98.9 & 96.8 & 88.1 & 4.3
& 100.0 & 93.0 & 73.5 \\

HeSS (CVPR 2026)
& 97.5 & 93.9 & 73.9 & 18.4
& 98.9 & 96.9 & 86.5 & {{3.4}}
& 100.0 & 86.6 & 61.6 \\

FastVGGT (ICLR 2026)
& 100.0 & 96.4 & 79.2 & 14.2
& 99.0 & 99.0 & 84.1
& \underline{\underline{2.8}}
& 100.0 & 90.5 & 71.0 \\

SparseVGGT (CVPR 2026)
& 99.9 & 95.3 & 74.2 & 16.2
& 96.7 & 98.9 & 83.8
& \underline{{2.7}}
& 100.0 & 83.2 & 57.6 \\

\midrule

VGGT-Prime (Ours)
& 99.9 & 95.1 & 78.6
& \textbf{9.8}
& 99.0 & 99.2 & 87.2
& \textbf{1.9}
& 100.0 & 90.8 & 72.5 \\

\bottomrule
\end{tabular}
\end{adjustbox}
\caption{
Pose quality and runtime comparison. RRA, RTA, and AUC are computed at $30^\circ$. Time is measured in seconds.
}
\label{tab:app_pose}
\end{table*}
\begin{table*}[h!]
\centering
\small
\setlength{\tabcolsep}{1mm}
\renewcommand{\arraystretch}{1.12}

\begin{tabular}{l|ccc|cc|cc}
\toprule
\multirow{2}{*}{\textbf{Method}}
& \multicolumn{3}{c|}{\textbf{7-Scenes}}
& \multicolumn{2}{c|}{\textbf{Bonn}}
& \multicolumn{2}{c}{\textbf{Sintel}} \\
\cmidrule(lr){2-4}
\cmidrule(lr){5-6}
\cmidrule(lr){7-8}
&
\textbf{AbsRel $\downarrow$}
& \textbf{$\delta < 1.25$ $\uparrow$}
& \textbf{Time(s) $\downarrow$}
&
\textbf{AbsRel $\downarrow$}
& \textbf{$\delta < 1.25$ $\uparrow$}
&
\textbf{AbsRel $\downarrow$}
& \textbf{$\delta < 1.25$ $\uparrow$} \\
\midrule

VGGT (CVPR 2025)
& 0.271 & 0.958 & 38.1
& 0.109 & 0.884
& 0.592 & 0.583 \\

\midrule

Fast3R (CVPR 2025)
& 0.345 & 0.889 & 21.2
& 0.153 & 0.831
& 0.829 & 0.352 \\

TTT3R (ICLR 2026)
& 0.314 & 0.949 & \underline{11.3}
& 0.110 & 0.886
& 0.717 & 0.464 \\

HTTM (CVPR 2026)
& 0.276 & 0.959 & \underline{\underline{12.1}}
& 0.118 & 0.859
& 0.647 & 0.538\\

HeSS (CVPR 2026)
& 0.277 & 0.962 & 18.4
& 0.105 & 0.901
& 0.577 & 0.520 \\

FastVGGT (ICLR 2026)
& 0.272 & 0.960 & 14.2
& 0.101 & 0.899
& 0.620 & 0.526 \\

SparseVGGT (CVPR 2026)
& 0.274 & 0.962 & 16.2
& 0.100 & 0.914
& 0.675 & 0.518 \\

\midrule

VGGT-Prime (Ours)
& 0.275 & 0.965
& \textbf{9.8}
& 0.101 & 0.894
& 0.524 & 0.591 \\

\bottomrule
\end{tabular}

\caption{
Depth estimation runtime comparison. Time is measured in seconds.
}
\label{tab:app_depth}
\end{table*}

\paragraph{Pose Estimation Results.}
For pose estimation, VGGT-Prime achieves the lowest measured runtime while maintaining accuracy close to VGGT both 7-Scenes and CO3Dv2. Similar trends hold on RealEstate10K, where VGGT-Prime outperforms both FastVGGT and SparseVGGT and slightly trails VGGT in pose estimation quality. Among the newly efficient models, TTT3R achieves the highest RealEstate10K AUC (77.3 versus 72.5) but performs substantially worse on 7-Scenes, whereas HTTM better preserves pose accuracy but remains slower than VGGT-Prime. HeSS performs consistently worse than HTTM across the pose benchmarks. Overall, VGGT-Prime delivers competitive pose estimation performance while having the lowest measured runtime.

\paragraph{Depth Estimation Results.}
As shown in Table~\ref{tab:app_depth}, VGGT-Prime achieves the lowest measured runtime while closely matching or improving upon VGGT in depth accuracy. On 7-Scenes, it trails VGGT by only 0.004 AbsRel while achieving the highest $\delta<1.25$, and on Sintel, it substantially improves both metrics. Although it slightly trails the strongest efficient baselines on Bonn, it remains competitive overall. Among these models, TTT3R provides the lowest runtime after VGGT-Prime but performs worse in accuracy, while HTTM and HeSS better preserve depth quality but are slow. Overall, VGGT-Prime achieves the lowest measured runtime while outperforming most baselines in depth-estimation.

\paragraph{Comparison with Head-wise Acceleration Methods.}

Existing head-wise acceleration methods for visual geometry transformers, including HTTM and HeSS, have two key limitations. 
First, they reduce token redundancy within attention heads while leaving the underlying VGGT architecture unchanged. 
Second, they rely on fixed reduction policies. Specifically, HTTM applies predetermined token-merging ratios, whereas HeSS uses a static head-importance ranking. 
VGGT-Prime instead targets architectural redundancy by dynamically routing each head to an appropriate compute mode according to its input-dependent saliency.
As shown in Tables~\ref{tab:scannet_appendix},~\ref{tab:other_pc},~\ref{tab:app_pose}, and~\ref{tab:app_depth}, VGGT-Prime is faster on both short- and long-sequence inputs while maintaining comparable output quality, demonstrating its favourable performance.

\section{Additional Efficiency Analysis}
\label{app:saliency}
\subsection{Flops Analysis} We report floating-point operations (FLOPs) as a hardware- and implementation-agnostic measure of computational cost. FLOPs quantify the number of arithmetic operations required by a model and provide a standardized proxy for comparing computational efficiency. Unlike runtime, FLOPs are agnostic to the hardware used, kernel implementation, or system level optimization. Specifically, we use PyTorch's profiler to estimate the FLOPs of each model across different numbers of input views on the ScanNet50 dataset. 

\begin{figure}[t!]
    \centering
    \includegraphics[width=\linewidth]{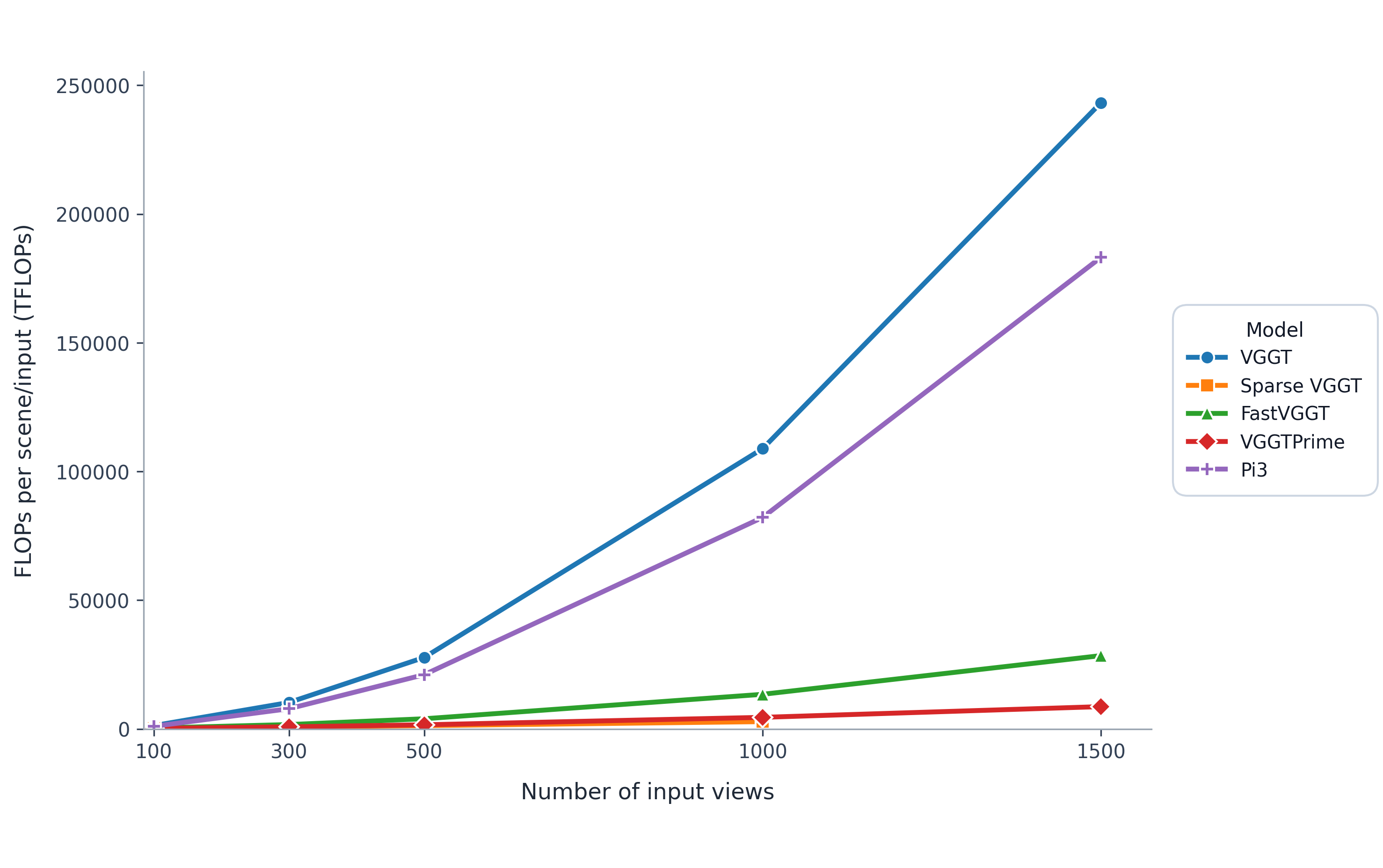}
    \caption{FLOPs on ScanNet50 across varying view counts.}
    \label{fig:dense_view_flops}
\end{figure}

Figure~\ref{fig:dense_view_flops} reports FLOPs on ScanNet50 as the number of input views increases. VGGT-Prime remains more computationally efficient than all evaluated methods. Moreover, the FLOP-based ranking remains stable across view counts and are broadly consistent with the wall-clock runtime results, indicating that both measures capture similar efficiency trends. One minor discrepancy is that SparseVGGT requires fewer FLOPs than FastVGGT and is comparable to VGGT-Prime in total FLOP count. However, this advantage does not translate into lower wall-clock latency because SparseVGGT relies on numerous small kernel launches, which introduce substantial execution overhead. SparseVGGT also encounters an out-of-memory error beyond 1K views. Overall, these results show that VGGT-Prime provides the strongest efficiency in both total computation and practical inference time.

\subsection{Theoretical Complexity of VGGT-Prime}
In this section, we analyze the theoretical time complexity of VGGT-Prime, focusing specifically on the global aggregator. We derive the Big-$\mathcal{O}$ complexity of VGGT-Prime by decomposing VGGT-Prime's compute into the fixed router cost and a variable mean pooling, surrogate, and full attention cost.

\paragraph{Preliminaries.} Let ($N$) denote the number of input tokens, ($d$) the per-head feature dimension, and ($H$) the total number of attention heads. We further denote ($H_{\mathrm{full}}$), ($H_{\mathrm{lin}}$), and ($H_{\mathrm{mean}}$) as the numbers of heads routed to full attention, the surrogate attention, and mean pooling, respectively.

\paragraph{Router.}
The router's cost consists of (i) attention-based pooling and (ii) single-query attention. The attention-based pooling step has complexity $\mathcal{O}(H N d^2)$, while single-query attention has complexity $\mathcal{O}(N d)$. Since the internal dimension $d=64$, the total router complexity across heads is:
\begin{equation}
\mathcal{O}(H N d^2).
\end{equation}

\paragraph{Compute Paths.}
For all heads, the dominant operator is materializing the attention operator. Hence for full attention we find: 
\begin{equation}
    \mathcal{O}(H_{\mathrm{full}} N^2 d).
\end{equation}
For heads routed to the surrogate attention, the cost scales linearly with the number of tokens:
\begin{equation}
    \mathcal{O}(H_{\mathrm{lin}} N d).
\end{equation}
For heads routed to mean pooling, the cost is also linear in the number of tokens:
\begin{equation}
    \mathcal{O}(H_{\mathrm{mean}} N d).
\end{equation}

\paragraph{Total.}
Combining these terms and assuming $d=64$, the total cost can be approximated as
\begin{equation}
\mathcal{O}
\left(
H_{\mathrm{full}} N^2 d + HN d^2
\right).
\end{equation}

Thus, the quadratic view-wise cost depends only on the number of heads routed to full attention. Under the chosen routing hyperparameter, more than 90\% of heads are routed to non-softmax attention paths. At this ratio, the model can practically process large sequences quickly, while maintaining competitive performance.
\begin{figure}[h!]
    \centering
    \includegraphics[width=\linewidth]{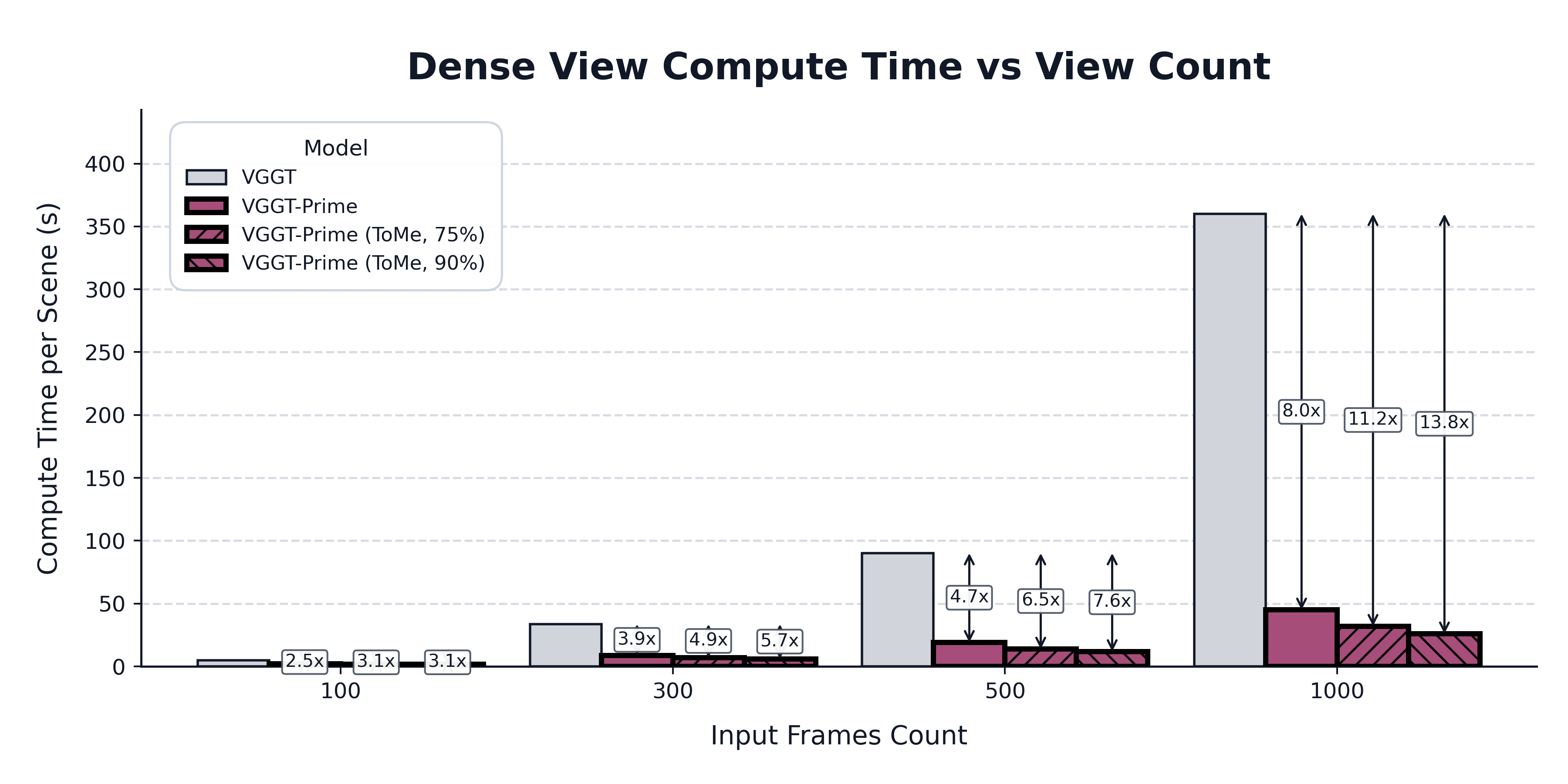}
    \caption{Full forward-pass compute time on ScanNet for VGGT, VGGT-Prime, and VGGT-Prime combined with token merging (ToMe) at merging ratios 75\% and 90\%.}
    \label{fig:ToMe}
\end{figure}

\section{Additional Experiments}
\label{app:experiments}
\subsection{VGGT-Prime \& Token Merging}

VGGT-Prime is designed as an architectural acceleration of VGGT. Hence, in principle, it can be combined with token reduction methods, such as the Token Merging (ToMe) used in FastVGGT. Figure~\ref{fig:ToMe} evaluates this combination. The resulting VGGT-Prime (ToMe) model improves the inference time at the 75\% token-merging ratio by $11\times$ and at 90\% token-merging by nearly $14\times$, showing the potential of combining architectural and token-space acceleration methods in terms of compute efficiency.

\subsection{Analysis of VGGT-Prime's Router}

To better understand whether VGGT-Prime’s router performs meaningful adaptive routing, we analyze its behavior from three complementary perspectives. 
First, we study how the choice of routing thresholds affects camera pose and depth estimation on CO3Dv2's validation split. 
Second, we examine whether the router avoids collapsing to a fixed strategy and instead makes input-dependent decisions. 
Finally, we assess whether the selected thresholds generalizes across datasets.

\paragraph{Determining Routing Thresholds.}
We begin by evaluating the router’s sensitivity to its two thresholds, $\tau_{\mathrm{low}}$ and $\tau_{\mathrm{high}}$. 
In particular, our goal is to optimize the thresholds for performance and runtime, such that the most full attention heads are removed without actually impacting performance. 
To assess this, we vary each threshold over low, medium, and high values while holding the other fixed, yielding five configurations in total. 
We evaluate these settings on a subset of CO3Dv2 validation scenes using pose and depth metrics.

\begin{table}[ht]
\centering
\small
\caption{
Sensitivity of VGGT-Prime to the routing thresholds on the CO3Dv2 validation subset.
The selected configuration is shown in bold.
}
\label{tab:threshold_sensitivity}
\setlength{\tabcolsep}{5pt}
\renewcommand{\arraystretch}{1.1}
\begin{tabular}{cc|ccc}
\toprule
$\boldsymbol{\tau_{\mathrm{low}}}$
& $\boldsymbol{\tau_{\mathrm{high}}}$
& \textbf{AUC@$30^\circ$ $\uparrow$}
& \textbf{AbsRel $\downarrow$}
& \textbf{Time (s) $\downarrow$} \\
\midrule
0.010 & 0.100 & 90.19 & 0.023 & 2.92 \\
0.010 & 0.500 & 58.78 & 0.145 & 1.92 \\
\midrule
0.005 & 0.200 & 88.16 & 0.026 & 2.23 \\
0.100  & 0.200 & 69.94 & 0.057 & 1.98 \\
\midrule
\textbf{0.010} & \textbf{0.200} & \textbf{87.24} & \textbf{0.026} & \textbf{2.04} \\
\bottomrule
\end{tabular}
\end{table}

Table~\ref{tab:threshold_sensitivity} shows that VGGT-Prime is most sensitive to $\tau_{\mathrm{high}}$. Increasing $\tau_{\mathrm{high}}$ beyond ($0.2$) substantially degrades pose and depth performance while yielding only a small runtime reduction, since moving a few highly salient heads moved away from full attention can cause a large loss in accuracy. In contrast, the model is less sensitive to $\tau_{\mathrm{low}}$, since increasing this threshold mainly shifts low-saliency heads from the surrogate attention to mean pooling. As previously, established these heads are mostly mean-pooling like, so the two approximations often produce similar outputs. 
We therefore use $\tau_{\mathrm{low}}=0.01$ and $\tau_{\mathrm{high}}=0.20$.

\begin{figure*}[t!]
    \centering
    \includegraphics[width=\linewidth]{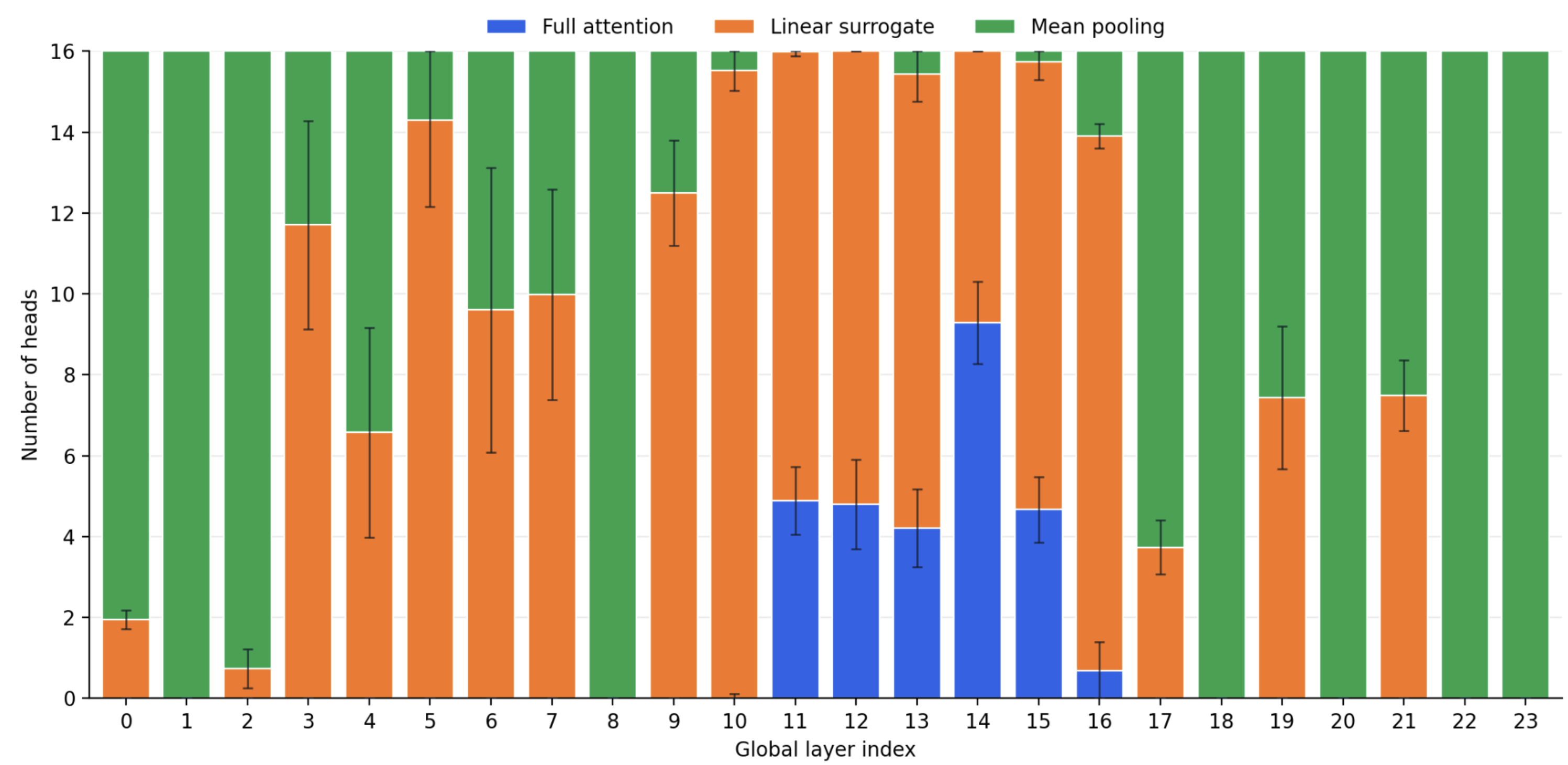}
    \caption{Distribution of routed attention heads across global attention layers on 7-scenes. Error bars are 1 standard deviation.}
    \label{fig:Layer_routing_distribution}
\end{figure*}

\paragraph{Layer Wise Routing Distribution.}

To determine whether the model is truly adaptive or simply learns a fixed routing policy, we analyze routing distributions across both layers and inputs. As shown in Figure~\ref{fig:Layer_routing_distribution}, the routing patterns show strong input-dependent variability across layers, indicating correct adaptive routing behavior. Early layers exhibit a mixture of mean pooling and surrogate attention with large input-dependent variability. The middle layers contain the highest proportion of linear-surrogate and full-attention heads. In contrast, the final layers are dominated by mean pooling with only a small number of linear-surrogate heads, suggesting that the computation required for global attention decreases toward the end of the network. Overall, these results show that routing remains heterogeneous across both layers and inputs, suggesting no layer wise mode collapse.

\begin{table}[t!]
\centering
\small
\caption{
Distribution of routing patterns across datasets.
}
\label{tab:router_distribution_datasets}
\begin{tabular}{lccc}
\toprule
\textbf{Dataset}
& \textbf{Mean Pooling}
& \textbf{Surrogate}
& \textbf{Full Attention} \\
\midrule
CO3Dv2     & 49.4\% & 43.2\% & 7.4\% \\
7-Scenes   & 49.3\% & 42.5\% & 8.2\% \\
ETH3D      & 48.8\% & 41.2\% & 9.9\% \\
Bonn       & 49.5\% & 42.5\% & 8.0\% \\
Re10K      & 51.2\% & 41.8\% & 7.0\% \\
ScanNet50  & 45.3\% & 45.8\% & 8.9\% \\
Sintel     & 55.4\% & 37.5\% & 7.1\% \\
\midrule
\textbf{Average}
& \textbf{50.0\%}
& \textbf{42.2\%}
& \textbf{7.8\%} \\
\bottomrule
\end{tabular}
\end{table}

\paragraph{Dataset Saliency Analysis.} 
Next, we examine whether the routing thresholds and underlying routing logic generalize across datasets. A key assumption in the main text is that a single pair of thresholds, $\tau_{\mathrm{low}}=0.01$ and $\tau_{\mathrm{high}}=0.2$, can be applied to all datasets. To evaluate this assumption, we compare the head routing distributions across datasets and show that they remain sufficiently consistent, providing empirical support for dataset-independent routing thresholds. 

As shown in Table~\ref{tab:router_distribution_datasets}, the head-routing distributions are highly consistent across datasets, with only minor variations. Mean pooling accounts for approximately 50\% of routed heads on most datasets, increasing to 55\% on Sintel and decreasing to 45\% on ScanNet. Conversely, the surrogate attention accounts for roughly 42\% on most datasets, with a lower proportion on Sintel at 38\% and a higher proportion on ScanNet at 46\%. Full attention remains consistently near 8\%, reaching 9\% on ScanNet. This likely stems from that fact that ScanNet contains scenes with occlusions and ambiguous cross-view correspondences that require more query-dependent attention, whereas Sintel’s consistent synthetic scenes permit greater use of mean pooling.

\begin{figure*}[t!]
    \centering
    \includegraphics[width=\linewidth]{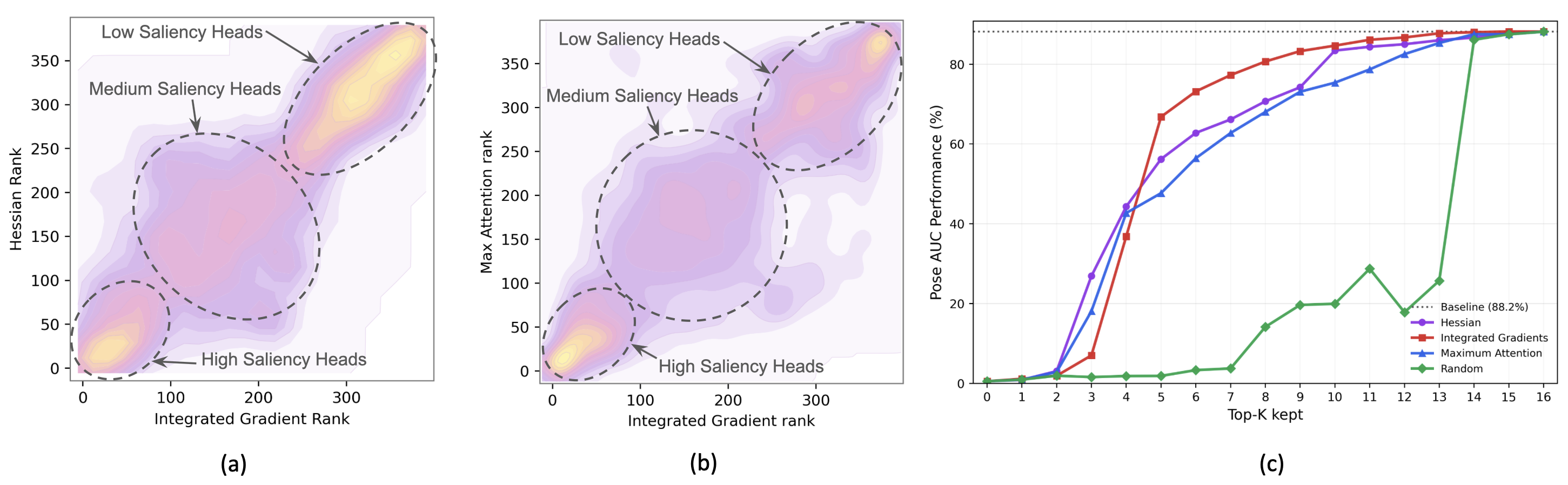}
    \caption{Head saliency analysis.
    (a/b) Correlation between head rankings determined by Hessian sensitivity score, maximum attention, and maximum integrated gradients. The rankings exhibit a significant Spearman's correlation coefficient of (i) $\rho = 0.741$ and (ii) $\rho = 0.852$.
    (c) Effect of pruning attention heads according to different saliency metrics on camera pose.}
    \label{fig:Head_Saliency}
\end{figure*}

\subsection{Hessian Head Saliency Metric}

In the main text, we introduced maximum integrated gradients (IG) and maximum attention as head saliency metrics. 
However, prior work by~\citet{Kim_2026_CVPR} showed that Hessian-based scores can also capture head importance suggesting that there exists multiple ways to calculate and define head-saliency. 
Here, we broaden the comparison to include the Hessian sensitivity score (HeSS) by re-implementing the metric to the best of our ability. We additionally, formally define the metrics introduced in the main text.

\paragraph{Preliminaries.} We begin by formally defining the saliency metrics introduced in the main paper.
Let \(O_h \in \mathbb{R}^{N \times d}\) denote the output of attention head \(h\), \(O_h^0\) the zero output baseline, and \(F\) the model's final feature representation. We define the maximum integrated-gradient saliency of head \(h\) as
\begin{equation}
    s_h^{\mathrm{IG}}
    =
    \max_{i,c}
    \left\|
    \operatorname{IG}\!\left(F | \{O_h, O_h^0\}\right)_{i,c}
    \right\|_1.
\end{equation}
Given the attention matrix \(A_h\), we define the maximum-attention saliency score as
\begin{equation}
    s_h^{\mathrm{attn}}
    =
    \max_{i,j} A_{h,i,j}.
\end{equation}
VGGT-Prime uses the raw saliency scores. 
However, for certain experiments, we rank score in descending order:
\begin{equation}
    r_h^{(m)}
    =
    1+\sum_{h'\neq h}
    \mathbb{I}\!\left[s_{h'}^{(m)}>s_h^{(m)}\right].
\end{equation}

\paragraph{Consistency with HeSS.}
Next, we validate whether maximum IG produces rankings consistent with HeSS~\cite{Kim_2026_CVPR}. As shown in Figure~\ref{fig:Head_Saliency}(a), the two metrics are strongly correlated, with a Spearman coefficient of $\rho=0.852$. We further examine whether the saliency clusters previously observed between maximum attention and IG also emerge with HeSS. Figure~\ref{fig:Head_Saliency}(b) provides the maximum-attention/IG comparison for reference. Both comparisons reveal three distinct clusters corresponding to low-, medium-, and high-saliency heads. These results suggest that the coarse saliency groups are not artifacts of the selected metrics and provide further support for the groupings discussed in the main text.

\paragraph{Head Pruning.} Lastly, we conduct a systematic study of the 2 head-saliency metrics discussed in this work along with HeSS. Specifically, we evaluate each metric by progressively removing heads according to their ranking and measuring the resulting degradation in model performance. 
As shown in Figure~\ref{fig:Head_Saliency}(c), all saliency-based methods outperform random head pruning, indicating that each captures a meaningful saliency signal. IG provides the strongest overall ranking, followed by HeSS and maximum attention.
Overall, these results validate that head saliency can be estimated reliably using several metrics, with IG providing the most robust ranking.

\section{Visualization of VGGT-Prime}
\label{app:Visual}

\begin{figure*}[h!]
    \centering
    \includegraphics[width=\linewidth]{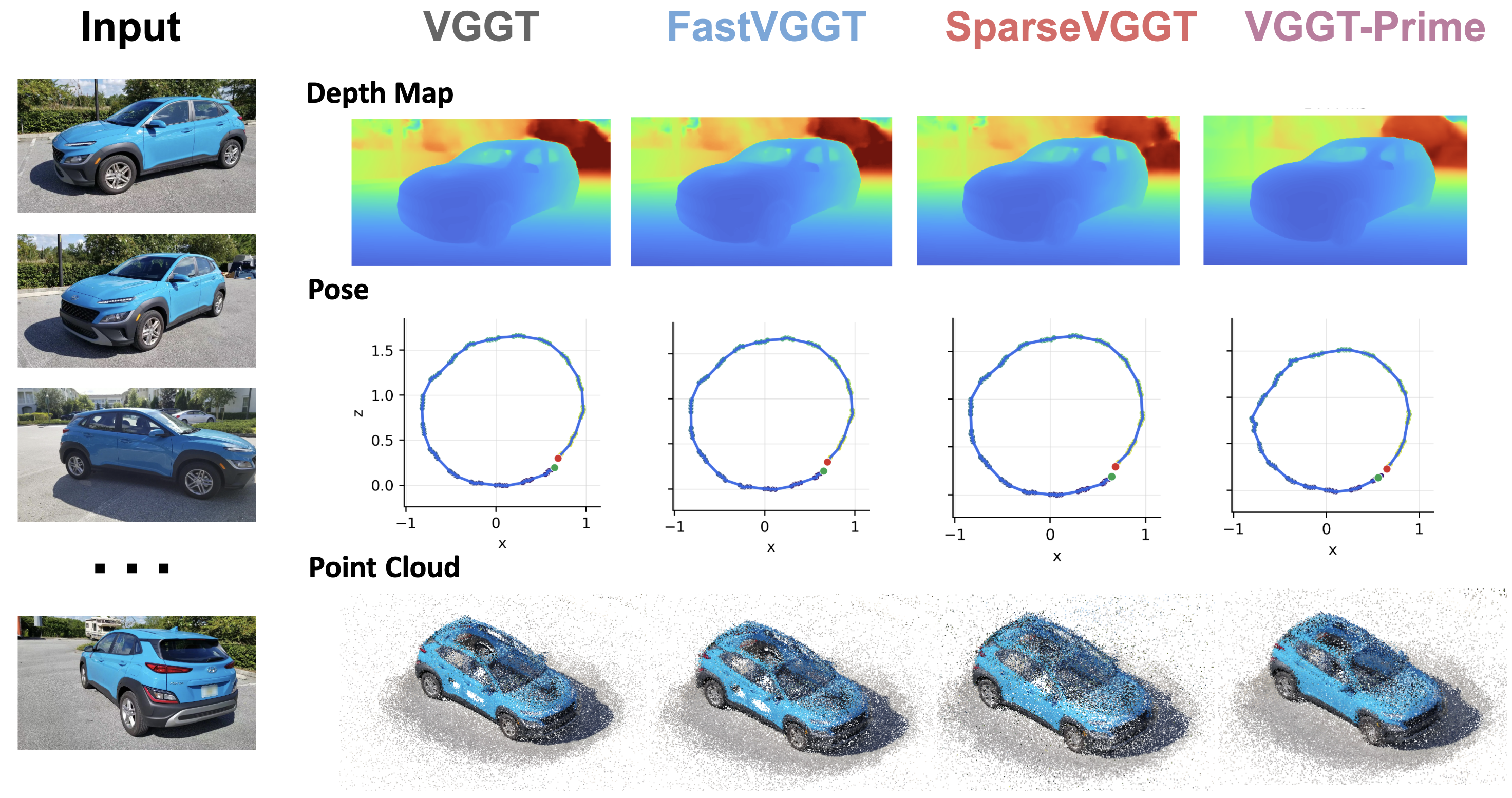}
    \caption{3D reconstruction of a blue vehicle in an outdoor CO3Dv2 scene. Input views in their correct positional order are shown on the left.}
    \label{fig:qualitative1}
\end{figure*}

\begin{figure*}[h!]
    \centering
    \includegraphics[width=\linewidth]{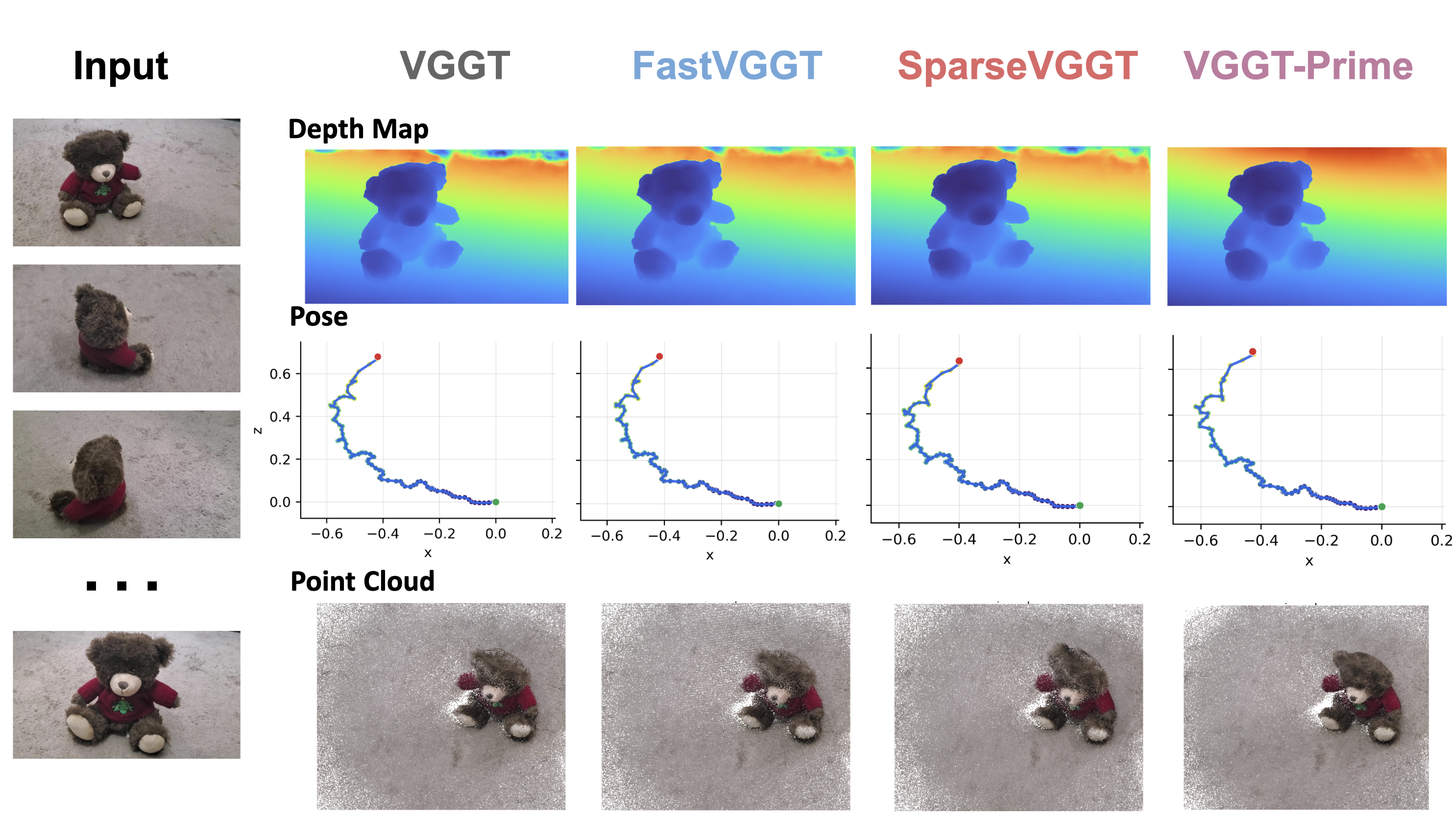}
    \caption{3D reconstruction of a stuffed animal in an indoor low light scene from CO3Dv2. Input views in their correct positional order are shown on the left.}
    \label{fig:qualitative2}
\end{figure*}

We provide representative qualitative comparisons of predicted depth maps, camera trajectories, and point clouds in Figures~\ref{fig:qualitative1} and~\ref{fig:qualitative2}, covering object-centric outdoor scenes and low texture indoor environments. For each scene, we process the first 100 frames with every model, and process all visualizations use identical plotting thresholds and viewpoints.

\paragraph{Object-Centric Outdoor Scene.}
As shown in Figure~\ref{fig:qualitative1}, VGGT-Prime produces a sharp depth map with a clear separation between the foreground vehicle and the surrounding scene. It also recovers a more complete point cloud than the competing methods, particularly around the vehicle door and interior which are relatively low texture regions.

\paragraph{Cluttered Indoor Scene.}
Figure~\ref{fig:qualitative2} shows that VGGT-Prime remains effective in a low texture indoor environment against a relatively smooth, constant color background. Its depth prediction preserves object boundaries against a relatively smooth carpet environment. The estimated trajectory also remains smooth and consistent with the other models. The resulting point cloud is more complete and better defined, especially around the bear's ears.

\end{document}